\documentclass[10pt,twocolumn,letterpaper]{article}

\usepackage{cvpr}
\usepackage{times}
\usepackage{epsfig}
\usepackage{graphicx}
\usepackage{amsmath}
\usepackage{amssymb}
\usepackage{pdfpages}
\usepackage{caption}
\usepackage{subcaption}
\usepackage{xcolor}

\usepackage[pagebackref=true,breaklinks=true,letterpaper=true,colorlinks,bookmarks=false]{hyperref}

\cvprfinalcopy 

\def\cvprPaperID{4628} 

\ifcvprfinal\fi
\begin{document}

\title{Selective Sensor Fusion for Neural Visual-Inertial Odometry}

\author{Changhao Chen\textsuperscript{1},
Stefano Rosa\textsuperscript{1},
Yishu Miao\textsuperscript{2},
Chris Xiaoxuan Lu\textsuperscript{1},
\\
Wei Wu\textsuperscript{3},
Andrew Markham\textsuperscript{1},
Niki Trigoni\textsuperscript{1}
\\
\textsuperscript{1}{Department of Computer Science, University of Oxford}\\
\textsuperscript{2}{MO Intelligence}
\textsuperscript{3}{Tencent}\\
}
\maketitle

\begin{abstract}
Deep learning approaches for Visual-Inertial Odometry (VIO) have proven successful, but they rarely focus on incorporating robust fusion strategies for dealing with imperfect input sensory data.
We propose a novel end-to-end selective sensor fusion framework for monocular VIO, which fuses monocular images and inertial measurements in order to estimate the trajectory whilst improving robustness to real-life issues, such as missing and corrupted data or bad sensor synchronization. 
In particular, we propose two fusion modalities based on different masking strategies: deterministic soft fusion and stochastic hard fusion, and we compare with previously proposed direct fusion baselines.
During testing, the network is able to selectively process the features of the available sensor modalities and produce a trajectory at scale.  We present a thorough investigation on the performances on three public autonomous driving, Micro Aerial Vehicle (MAV) and hand-held VIO datasets.
The results demonstrate the effectiveness of the fusion strategies, which offer better performances compared to direct fusion, particularly in presence of corrupted data.
In addition, we study the interpretability of the fusion networks by visualising the masking layers in different scenarios and with varying data corruption, revealing interesting correlations between the fusion networks and imperfect sensory input data.



\end{abstract}

\section{Introduction}

Humans are able to perceive their self-motion through space via multimodal perceptions. 
Optical flow (visual cues) and vestibular signals (inertial motion sense) are the two most sensitive cues for determining self-motion \cite{Fetsch2009}. 




In the fields of computer vision and robotics, integrating visual and inertial information in the form of Visual-Inertial Odometry (VIO) is a well researched topic \cite{jones2011visual,Li2013b,okvis,forster2017manifold,vinsmono}, as it enables ubiquitous mobility for mobile agents by providing robust and accurate pose information. Moreover, cameras and inertial sensors are relatively low-cost, power-efficient and widely found in ground robots, smartphones, and unmanned aerial vehicles (UAVs). 
Existing VIO approaches generally follow a standard pipeline that involves fine-tuning of both feature detection and tracking, and of the sensor fusion strategy. These models rely on handcrafted features, and fuse the information based on filtering \cite{Li2013b} or nonlinear optimization \cite{okvis,forster2017manifold,vinsmono}. However, naively using all features before fusion will lead to unreliable state estimation, as incorrect feature extraction or matching cripples the entire system. Real issues which can and do occur include camera occlusion or operation in low-light conditions \cite{Yang2018}, excess noise or drift within the inertial sensor \cite{NaserEl-SheimyHaiyingHou2008}, time-synchronization between the two streams or spatial misalignment \cite{Ling2018}. 

    \begin{figure*}
     	\centering
         \includegraphics[width=0.8\textwidth]{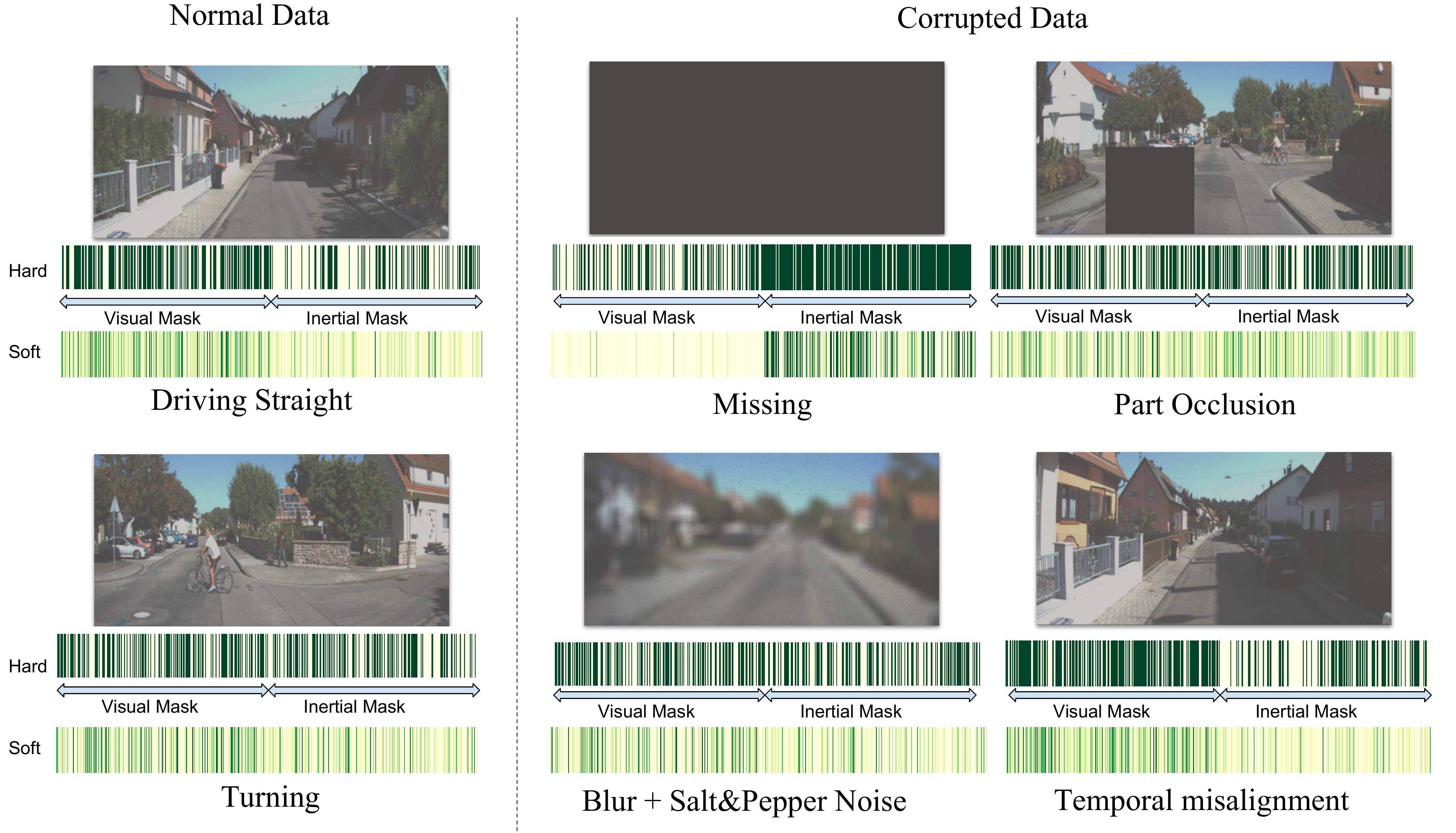}
         \caption{Visualization of the learned hard and soft fusion masks under different conditions (left: normal data; middle and right: corrupted data). The number (hard) or weights (soft) of selected features in the visual and inertial sides can reflect the self-motion dynamics (increasing importance of inertial features during turning), and data corruption conditions. 
         }
         \label{fig:mask}
     \end{figure*}

Recent studies on applying deep neural networks (DNNs) to solving visual-inertial odometry \cite{clarkwang2017} or visual odometry \cite{kendall2015posenet,Clark2017} showed competitive performance in terms of both accuracy and robustness. Although DNNs excel at extracting high-level features representative of egomotion, these learning-based methods are not explicitly modelling the sources of degradation in real-world usages. Without considering possible sensor errors, all features are directly fed into other modules for further pose regression in \cite{Brahmbhatt2018,Clark2017,kendall2015posenet}, or simply concatenated as in \cite{clarkwang2017}. These factors can possibly cause troubles to the accuracy and safety of VIO systems, when the input data are corrupted or missing.

For this reason, we present a generic framework that models feature selection for robust sensor fusion. The selection process is conditioned on the measurement reliability and the dynamics of both egomotion and environment. Two alternative feature weighting strategies are presented: soft fusion, implemented in a deterministic fashion; and hard fusion, which introduces stochastic noise and intuitively learns to keep the most relevant feature representations, while discarding useless or misleading information. 
Both architectures are trained in an end-to-end fashion.

By explicitly modelling the selection process, we are able to demonstrate the strong correlation between the selected features and the environmental/measurement dynamics by visualizing the sensor fusion masks, as illustrated in Figure \ref{fig:mask}. Our results show that features extracted from different modalities (i.e., vision and inertial motion) are complementary in various conditions: the inertial features contribute more in presence of fast rotation, while visual features are preferred during large translations (Figure \ref{fig:correlation}). Thus, the selective sensor fusion provides insight into the underlying strengths of each sensor modality. We also demonstrate how incorporating selective sensor fusion makes VIO robust to data corruption typically encountered in real-world scenarios. 

The main contributions of this work are as follows:
\begin{itemize}
    \item We present a generic framework to learn selective sensor fusion enabling more robust and accurate ego-motion estimation in real world scenarios.
    \item Our selective sensor fusion masks can be visualized and interpreted, providing deeper insight into the relative strengths of each stream, and guiding further system design.
    \item We create challenging datasets on top of current public VIO datasets by considering seven different sources of sensor degradation, and conduct a new and complete study on the accuracy and robustness of deep sensor fusion in presence of corrupted data.
\end{itemize}


	\begin{figure*}
     	\centering
         \includegraphics[width=0.8\textwidth]{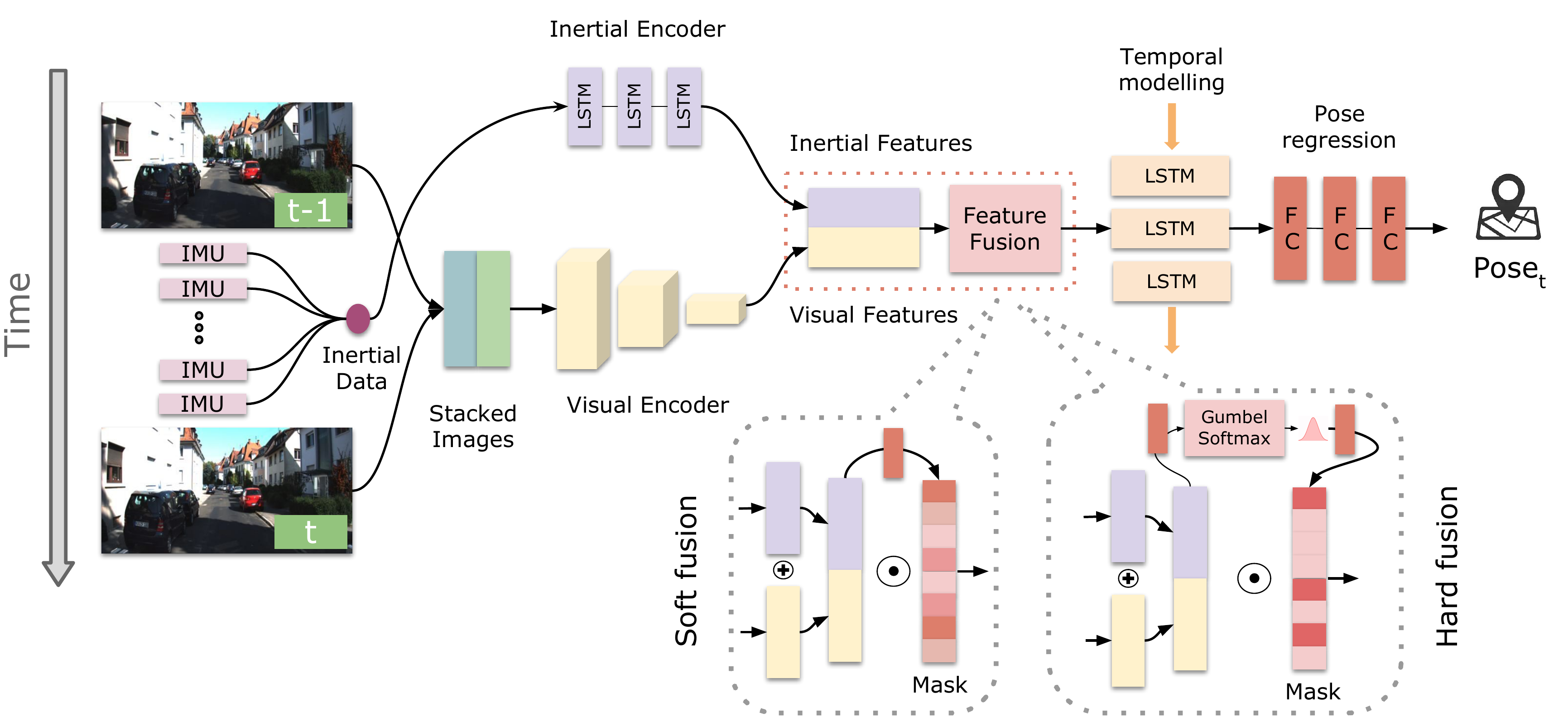}
         \caption{An overview of our neural visual-inertial odometry architecture with proposed selective sensor fusion, consisting of visual and inertial encoders, feature fusion, temporal modelling and pose regression. In the feature fusion component, we compare our proposed soft and hard selective sensor fusion strategies with direct fusion. 
         }
         \label{fig:framework}
     \end{figure*}

\vspace{-0.3cm}
\section{Neural VIO Models with Selective Fusion}
In this section, we introduce the end-to-end architecture for neural visual-inertial odometry, which is the foundation for our proposed framework. Figure \ref{fig:framework}(top) shows a modular overview of the architecture, consisting of visual and inertial encoders, feature fusion, temporal modelling and pose regression. Our model takes in a sequence of raw images and IMU measurements, and generates their corresponding pose transformation. With the exception of our novel feature fusion, the pipeline can be any generic deep VIO technique.
In the Feature Fusion component we propose two different selection mechanisms (soft and hard) and compare them with direct (i.e. a uniform/unweighted mask) fusion, as shown in Figure \ref{fig:framework}(bottom). 

\subsection{Feature Encoder}
\noindent \textbf{Visual Feature Encoder}
The Visual Encoder extracts a latent representation from a set of two consecutive monocular images $\mathbf{x}_V$. 
Ideally, we want the Visual Encoder $f_{\text{vision}}$ to learn geometrically meaningful features rather than features related with appearance or context. 
For this reason, instead of using a PoseNet model \cite{kendall2015posenet}, as commonly found in other DL-based VO approaches \cite{zhou2017unsupervised,Zhan2018,Yin2018}, we use FlowNetSimple \cite{Fischer2015} as our feature encoder.
Flownet provides features that are suited for optical flow prediction. 
The network consists of nine convolutional layers. The size of the receptive fields gradually reduces from 7$\times$7 to 5$\times$5 and finally 3$\times$3, with stride two for the first six.
Each layer is followed by a ReLU nonlinearity except for the last one, and we use the features from the last convolutional layer
 $\mathbf{a}_V$ as our visual feature:
\begin{equation}
    \mathbf{a}_V = f_{\text{vision}}(\mathbf{x}_V).
\end{equation}

\noindent \textbf{Inertial Feature Encoder:}
Inertial data streams have a strong temporal component, and are generally available at higher frequency ($\sim$100 Hz) than images ($\sim$10 Hz). Inspired by IONet \cite{ionet2018}, we use a two-layer Bi-directional LSTM with 128 hidden states as the Inertial Feature Encoder $f_{\text{inertial}}$. As shown in Figure \ref{fig:framework}, a window of inertial measurements $\mathbf{x}_I$ between each two images is fed to the inertial feature encoder in order to extract the dimensional feature vector ${\mathbf{a}_I}$:
\begin{equation}
    \mathbf{a}_I = f_\text{inertial}(\mathbf{x}_I).
\end{equation}

\subsection{Fusion Function}
We now combine the high-level features produced by the two encoders from raw data sequences, with a fusion function $g$ that combines information from the visual $\mathbf{a}_V$ and inertial $\mathbf{a}_I$ channels to extract the useful combined feature $\mathbf{z}$ for future pose regression task:
    \begin{equation}
        \mathbf{z} = g(\mathbf{a}_V, \mathbf{a}_I).
    \end{equation}
There are several different ways to implement this fusion function.
The current approach is to directly concatenate the two features together into one feature space (we call this method direct fusion $g_\text{direct}$). However, in order to learn a robust sensor fusion model, we propose two fusion schemes -- deterministic soft fusion $g_\text{soft}$ and stochastic hard fusion $g_\text{hard}$, which explicitly model the feature selection process according to the current environment dynamics and the reliability of the data input. Our selective fusion mechanisms re-weights the concatenated inertial-visual features, guided by the concatenated features themselves. The fusion network is another deep neural network and is end-to-end trainable. Details will be discussed in Section~\ref{sec:selective}.

\subsection{Temporal Modelling and Pose Regression}

The fundamental tenet of ego-motion estimation requires modeling temporal dependencies to derive accurate pose regression. Hence, a recurrent neural network (a two-layer Bi-directional LSTM) takes in input the combined feature representation $\mathbf{z}_{t}$ at time step $t$ and its previous hidden states $\mathbf{h}_{t-1}$ and models the dynamics and connections between a sequence of features. 
After the recurrent network, a fully-connected layer serves as the pose regressor, mapping the features to a pose transformation $\mathbf{y}_t$, representing the motion transformation over a time window. 
    \begin{equation}
        \mathbf{y}_t = \mathbf{RNN}(\mathbf{z}_{t}, \mathbf{h}_{t-1})
    \end{equation}

\section{Selective Sensor Fusion}
\label{sec:selective}
Intuitively, the features from each modality offer different strengths for the task of regressing pose transformations. 
This is particularly true in the case of visual-inertial odometry (VIO), where the monocular visual input is capable of estimating the appearance and geometry of a 3D scene, but is unable to determine the metric scale \cite{forster2017manifold}.
Moreover, changes in illumination, textureless areas and motion blur can lead to bad data association.
Meanwhile, inertial data is interoceptive/egocentric and generally environment-agnostic, and can still be reliable when visual tracking fails \cite{ionet2018}. 
However, measurements from low-cost MEMS inertial sensors are corrupted by inevitable noise and bias, which leads to higher long-term drift than a well-functioning visual-odometry chain.

Our perspective is that simply considering all features as though they are correct, without any consideration of degradation, is unwise and will lead to unrecoverable errors. In this section, we propose two different selective sensor fusion schemes for explicitly learning the feature selection process: soft (deterministic) fusion, and hard (stochastic) fusion, as illustrated in Figure \ref{fig:feature_fusion}. In addition, we also present a straightforward sensor fusion scheme -- direct fusion -- as a baseline model for comparison.

\subsection{Direct Fusion}
A straightforward approach for implementing sensor fusion in a VIO framework consists in the use of Multi-Layer Perceptrons (MLPs) to combine the features from the visual and inertial channels. 
Ideally, the system learns to perform feature selection and prediction in an end-to-end fashion.
Hence, direct fusion is modelled as:
\begin{equation}
    g_\text{direct}(\mathbf{a}_V, \mathbf{a}_I) = [\mathbf{a}_V; \mathbf{a}_I]
\end{equation}
where $[\mathbf{a}_V; \mathbf{a}_I]$ denotes an MLP function that concatenates $\mathbf{a}_V$ and $\mathbf{a}_I$.
                                
\subsection{Soft Fusion (Deterministic)}

We now propose a soft fusion scheme that explicitly and deterministically models feature selection.
Similar to the widely applied attention mechanism \cite{Vaswani2017,Xu2015,Hori2017}, this function re-weights each feature by conditioning on both the visual and inertial channels, which allows the feature selection process to be jointly trained with other modules. The function is deterministic and differentiable.

Here, a pair of continuous masks $\mathbf{s_V}$ and $\mathbf{s_I}$ is introduced to implement soft selection of the extracted feature representations, before these features are passed to temporal modelling and pose regression:
    \begin{eqnarray}
        &\mathbf{s}_V = \text{Sigmoid}_V([\mathbf{a}_V; \mathbf{a}_I]) \\
        &\mathbf{s}_I = \text{Sigmoid}_I([\mathbf{a}_V; \mathbf{a}_I])
    \end{eqnarray}
where $\mathbf{s_V}$ and $\mathbf{s_I}$ are the masks applied to visual features and inertial features respectively, and which are deterministically parameterised by the neural networks, conditioned on both the visual $\mathbf{a}_V$ and inertial features $\mathbf{a}_I$.
The sigmoid function makes sure that each of the features will be re-weighted in the range $[0,1]$.

Then, the visual and inertial features are element-wise multiplied with their corresponding soft masks as the new re-weighted vectors. The selective soft fusion function is modelled as
\begin{equation}
    g_\text{soft}(\mathbf{a}_V,\mathbf{a}_I)= 　[\mathbf{a}_V \odot \mathbf{s}_V;  \mathbf{a}_I \odot \mathbf{s}_I] .
\end{equation}





\subsection{Hard Fusion (Stochastic)}

    \begin{figure}[t]
     	\centering
         \includegraphics[width=0.85\columnwidth]{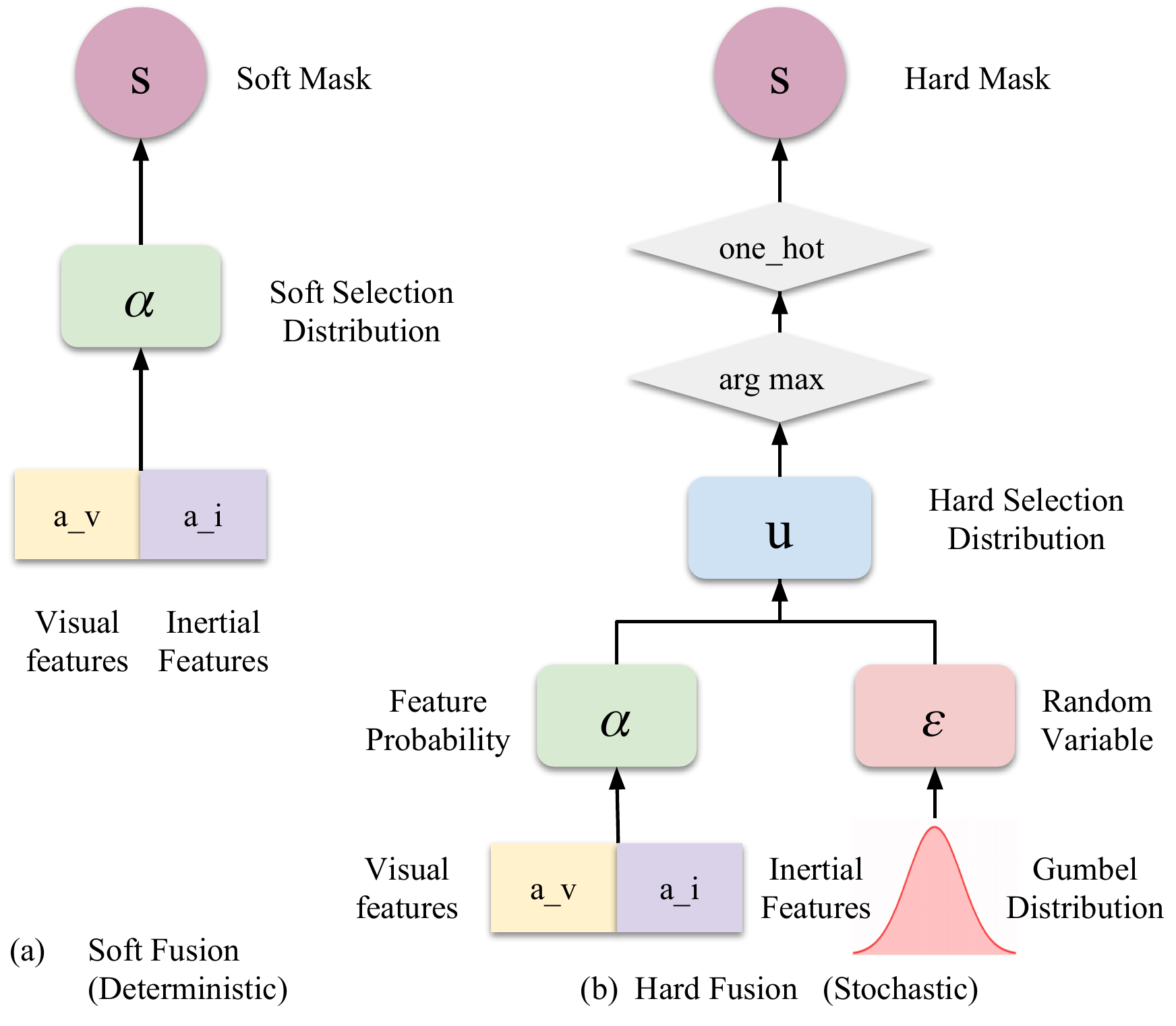}
         \caption{An illustration of our proposed soft (deterministic) and hard (stochastic) feature selection process.}
         \label{fig:feature_fusion}
     \end{figure}

In addition to the soft fusion introduced above, we propose a variant of the fusion scheme -- hard fusion.
Instead of re-weighting each feature by a continuous value, hard fusion learns a stochastic function that generates a binary mask that either propagates the feature or blocks it.
This mechanism can be viewed as a switcher for each component of the feature map, which is stochastic neural implemented by a parameterised Bernoulli distributions. 

However, the stochastic layer cannot be trained directly by back-propagation, as gradients will not propagate through discrete latent variables. 
To tackle this, the REINFORCE algorithm \cite{williams1992simple,mnih2014neural} is generally used to construct the gradient estimator.
In our case, we employ a more lightweight method -- Gumbel-Softmax resampling \cite{jang2016categorical,maddison2016concrete} to infer the stochastic layer, so that the hard fusion can be trained in an end-to-end fashion as well.

Instead of learning masks deterministically from features, hard masks $\mathbf{s}_V$ and $\mathbf{s}_I$ are re-sampled from a Bernoulli distribution, parameterised by $\mathbf{\alpha}$, which is conditioned on features but with the addition of stochastic noise:
    \begin{eqnarray}
       &\mathbf{s}_V \sim  p(\mathbf{s}_V | \mathbf{a}_V,\mathbf{a}_I) = \text{Bernoulli}(\mathbf{\alpha}_V) \\
       &\mathbf{s}_I \sim p(\mathbf{s}_I | \mathbf{a}_V,\mathbf{a}_I) = \text{Bernoulli}(\mathbf{\alpha}_I).
    \end{eqnarray}
Similar to soft fusion, features are element-wise multiplied with their corresponding hard masks as the new reweighted vectors. The stochastic hard fusion function is modelled as
    \begin{equation}
        g_\text{hard}(\mathbf{a}_V,\mathbf{a}_I)= [\mathbf{a}_V \odot \mathbf{s}_V; \mathbf{a}_I \odot \mathbf{s}_I]. 
    \end{equation}
Figure \ref{fig:feature_fusion} (b) shows the detailed workflow of proposed Gumbel-Softmax resampling based hard fusion. A pair of probability variables $\mathbf{\alpha}_V$ and $\mathbf{\alpha}_I$ is conditioned on the concatenated visual and inertial feature vectors $[\bf{a}_V; \bf{a}_I]$:
    \begin{eqnarray}
       &\mathbf{\alpha}_V = \text{Sigmoid}_V([\bf{a}_V; \bf{a}_I]) \\
       &\mathbf{\alpha}_I = \text{Sigmoid}_I([\bf{a}_V; \bf{a}_I]),
    \end{eqnarray}
where the probability variables are n-dimensional vectors $\mathbf{\alpha} = [\pi_1, ..., \pi_n ]$, representing the probability of each feature at location $n$ to be selected or not. Sigmoid function enables each vector to be re-weighted in the range $[0,1]$.

The Gumbel-max trick \cite{Maddison2014} allows efficiently to draw samples $\mathbf{s}$ from a categorical distribution given the class probabilities $\mathbf{\pi}_i$ and a random variable $\mathbf{\epsilon}_i$, and then the one-hot encoding performs "binarization" of the category:
    \begin{equation}
        \label{eq:s resample}
        \mathbf{s} = \text{one\_hot} (\arg \max_{\substack{i}} [\epsilon_i+\log \pi_i]).
    \end{equation}
This is due to the fact that for any $B \subseteq [1,...,n]$ \cite{Gumbel1954}:
    \begin{equation}
        \arg \max_{\substack{i}} [\epsilon_i+\log \pi_i] \sim \frac{\pi_i}{\sum_{i \in B} \pi_i}
    \end{equation}
It could be viewed as a process of adding independent Gumbel perturbations $\epsilon_i$ to the discrete probability variable. In practice, the random variable $\epsilon_i$ is sampled from a Gumbel distribution, which is a continuous distribution on the simplex that can approximate categorical samples:
\begin{equation}
    \epsilon = -\log(-\log(u)), u \sim \text{Uniform}(0,1).
\end{equation}
In Equation \ref{eq:s resample} the argmax operation is not differentiable, so Softmax function is instead used as an approximate:
    \begin{equation}
        h_i = \frac{\exp((\log (\pi_i)+\epsilon_i)/\tau)}{\sum_{i=1}^{n} \exp((\log (\pi_j)+\epsilon_j)/\tau)}, i=1,...,n,
    \end{equation}
where $\tau > 0$ is the temperature that modulates the re-sampling process.

\subsection{Discussions on Neural and classical VIOs}

Basically, soft fusion gently re-weights each feature in a deterministic way, while hard fusion directly blocks features according to the environment and its reliability.
In general, soft fusion is a simple extension of direct fusion that is good for dealing with the uncertainties in the input sensory data.
By comparison, the inference in hard fusion is more difficult, but it offers a more intuitive representation. The stochasticity gives the VIO system better generalisation ability and higher tolerance to imperfect sensory data.
The stochastic mask of hard fusion acts as an inductive bias, separating the feature selection process from prediction, which can also be easily interpreted by corresponding to uncertainties of the input sensory data.

Filtering methods update their belief based on the past state and current observations of visual and inertial modalities \cite{Mourikis2007,Li2013b,Hu2014,rovio}. "Learning" within these methods is usually constrained to gain and covariances \cite{Bishop2006}. This is a deterministic process, and noise parameters are hand-tuned beforehand.
Deep leaning methods are instead fully learned from data and the hidden recurrent state only contains information relevant to the regressor. Our approach models the feature selection process explicitly with the use of soft and hard masks. Loosely, the proposed soft mask can be viewed as similar to tuning the gain and covariance matrix in classical filtering methods, but based on the latent data representation instead.


\section{Experiments}
We evaluate our proposed approaches on three well-known datasets: the KITTI Odometry dataset for autonomous driving \cite{Geiger2013}, the EuRoC dataset for micro aerial vehicle \cite{euroc}, and the PennCOSYVIO dataset for hand-held devices \cite{penncosyvio}. A demonstration video and other details can be found at our \href{https://changhaoc.github.io/selective\_sensor\_fusion/}{project website} \footnote{https://changhaoc.github.io/selective\_sensor\_fusion/}. 

	 \begin{figure}
    	\centering
        \begin{subfigure}[t]{0.23\textwidth}
        	\includegraphics[width=\textwidth]{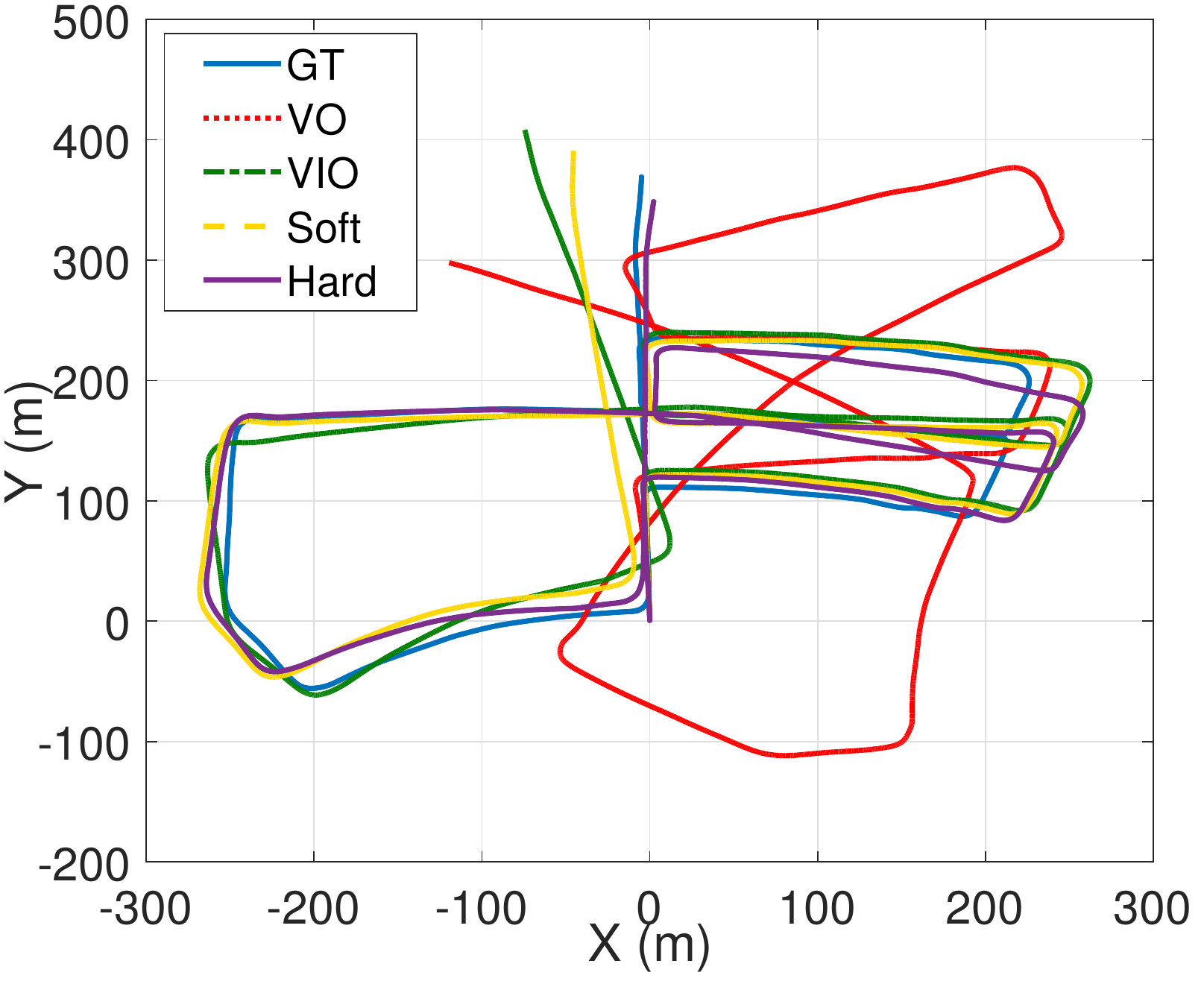}
        	\caption{\label{fig:traj_vicon} Seq 05 with vision degrad.}
        \end{subfigure}
        \begin{subfigure}[t]{0.23\textwidth}
        	\includegraphics[width=\textwidth]{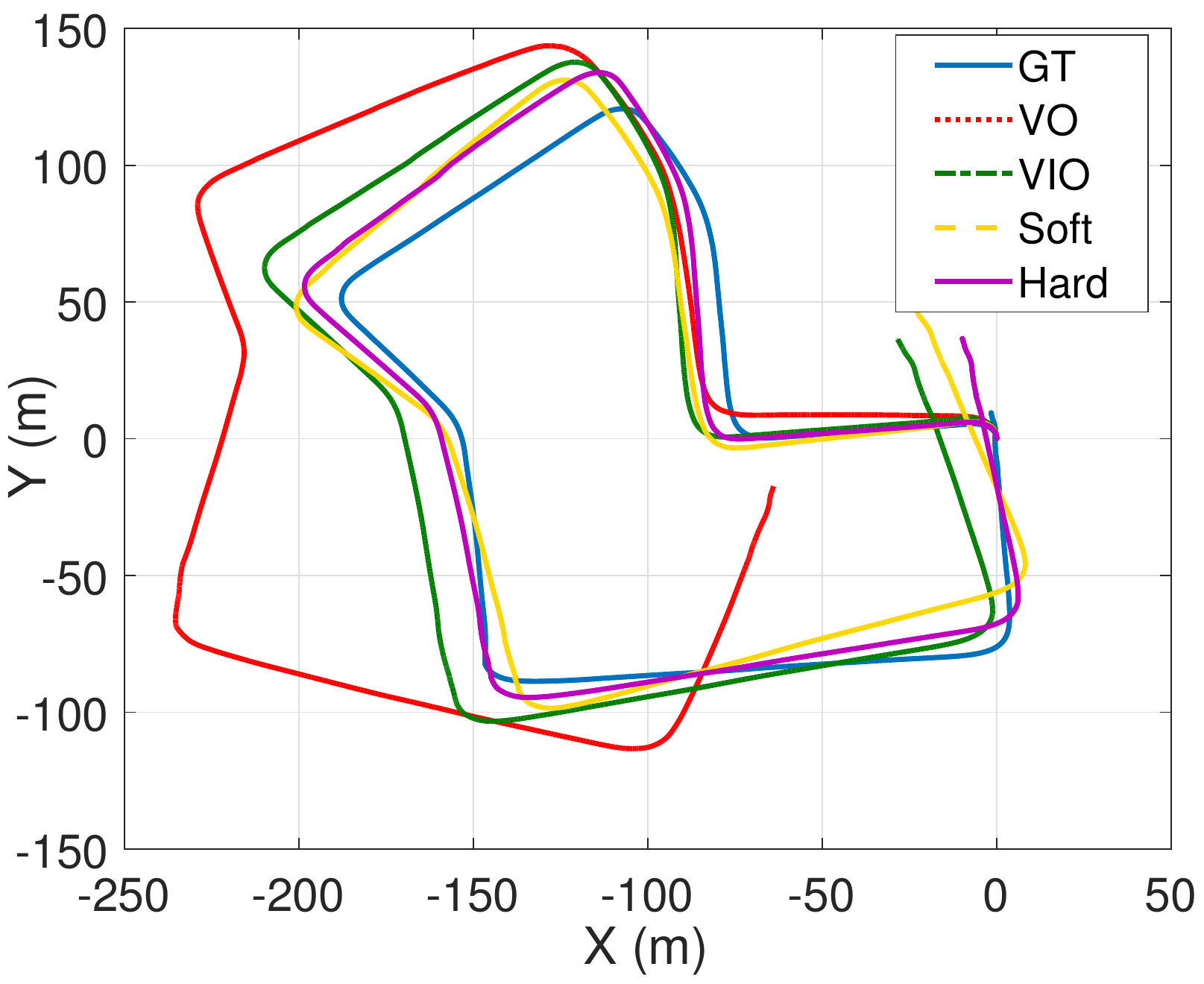}
        	\caption{\label{fig:traj_ionet} Seq 07 with vision degrad.}
        \end{subfigure}
         \begin{subfigure}[t]{0.23\textwidth}
        	\includegraphics[width=\textwidth]{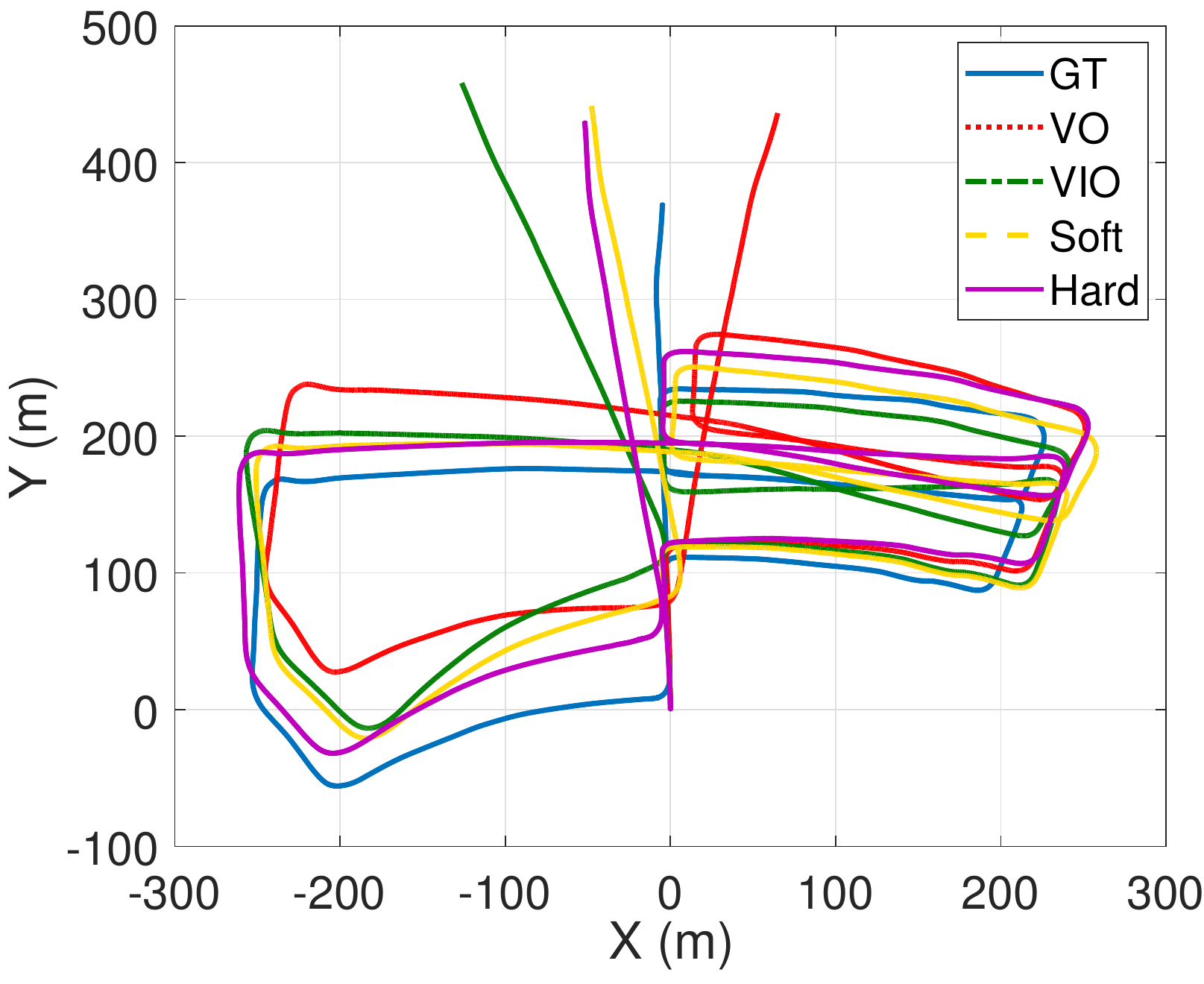}
        	\caption{\label{fig:traj_vicon} Seq 05 with all degradation}
        \end{subfigure}
        \begin{subfigure}[t]{0.23\textwidth}
        	\includegraphics[width=\textwidth]{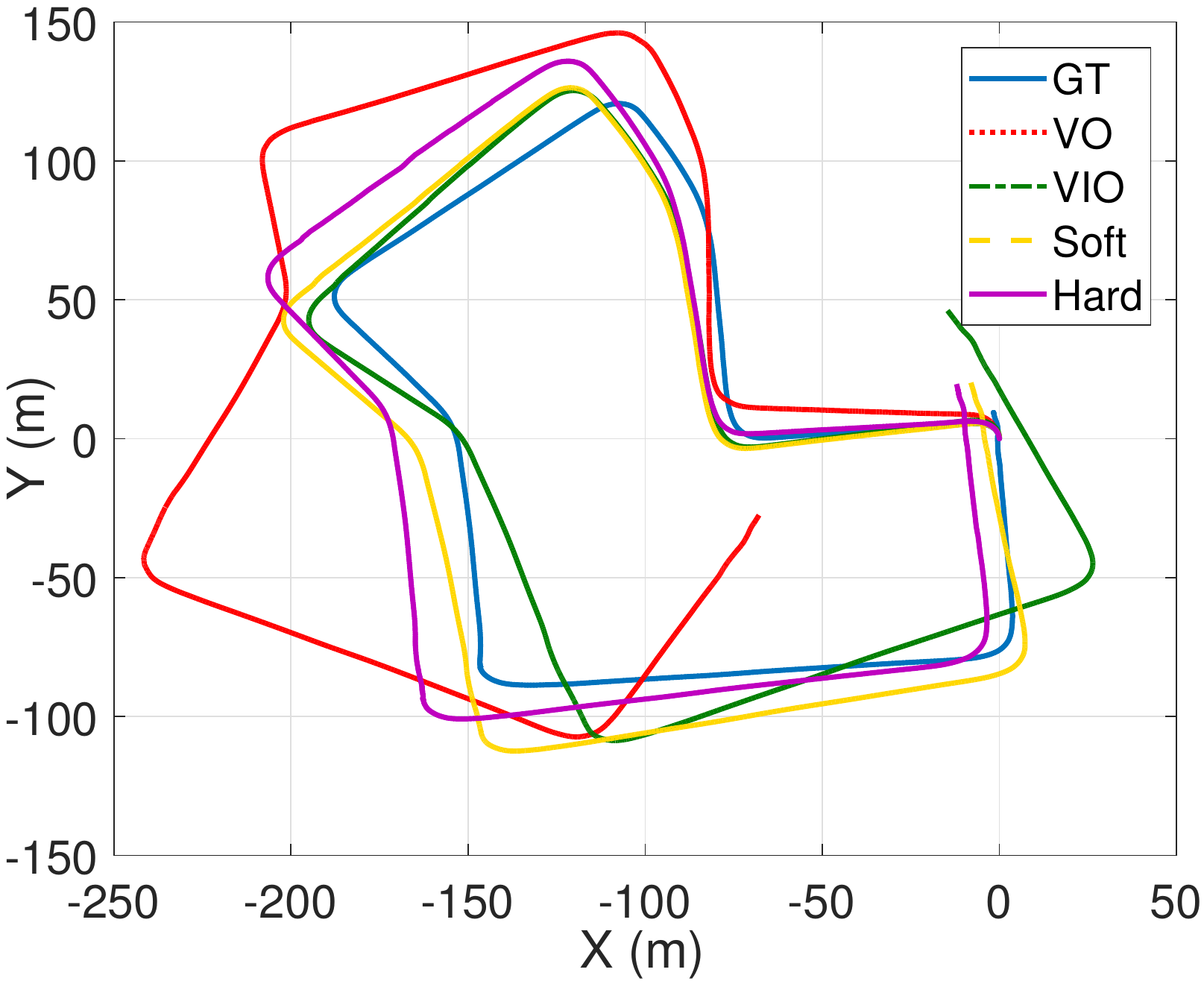}
        	\caption{\label{fig:traj_ionet} Seq 07 with all degradation}
        \end{subfigure}
        \caption{\label{fig:trajectories} Estimated trajectories on the KITTI dataset. Top row: dataset with vision degradation (10\% occlussion, 10\% blur, and 10\% missing data); bottom row: data with all degradation (5\% for each). Here, GT, VO, VIO, Soft and Hard mean the ground truth, neural vision-only model, neural visual inertial models with direct, soft, and hard fusion.} 
    \end{figure} 
    
    	\begin{table*}[t]
  		\caption{Effectiveness of different sensor fusion strategies in presence of different kinds of sensor data corruption. For each case we report absolute translational error (m) and rotational error (degrees).}
  		\label{tab:kittiseparatestudy}
  		\centering
  		\begin{tabular}{c|c|c|c|c|c|c|c}
    		        	    & \multicolumn{3}{c|}{Vision Degradation}
    		& \multicolumn{2}{c|}{IMU Degradation} & \multicolumn{2}{c}{Sensor Degradation} \\
    		Model & Occlusion & Blur & Missing & Noise and bias & Missing & Spatial & Temporal \\
    		\hline
    		Vision Only&0.117,0.148& 0.117,0.153&0.213,0.456&0.116,0.136&0.116,0.136&0.116,0.136& 0.116,0.136\\
            VIO Direct&0.116,0.110&0.117,0.107&0.191,0.155&0.118,0.115&0.118,0.163&0.119,0.137& 0.120,0.111\\
            VIO Soft&0.116,\textbf{0.105}&0.119,\textbf{0.104}&0.198,\textbf{0.149}&0.119, \textbf{0.105}&0.118,\textbf{0.129}&0.119,\textbf{0.128}& 0.119,\textbf{0.108}\\
            VIO Hard&\textbf{0.112},0.126&\textbf{0.114},0.110&\textbf{0.187},0.159& \textbf{0.114},0.120&\textbf{0.115},0.140&\textbf{0.111},0.146&\textbf{0.113},0.133\\
  		\end{tabular}
	\end{table*}

\subsection{Experimental Setup and Baselines}
The architecture was implemented with PyTorch and trained on a NVIDIA Titan X GPU. 

We chose the neural vision-only model and the neural visual-inertial model with direct fusion as our baselines, termed Vision-Only (DeepVO) and VIO-Direct (VINet) respectively in our experiments. The neural vision-only model uses the visual encoder, temporal modelling and pose regression as in our proposed framework in Figure \ref{fig:framework}. Neural visual inertial model with direct fusion uses the same framework as in our proposed selective fusion except the feature fusion component.
All of the networks including baselines were trained with a batch size of 8 using the Adam optimizer, with a learning rate $\text{lr}=1e^{-4}$. The hyper-parameters inside the networks were identical for a fair comparison.

\subsection{Datasets}
\textbf{KITTI Odometry dataset} \cite{Geiger2013} We used Sequences \textit{00, 01, 02, 04, 06, 08, 09} for training and tested the network on Sequences \textit{05}, \textit{07}, and \textit{10}, excluding sequence 03 as the corresponding raw file is unavailable. The images and ground-truth provided by GPS are collected at 10 Hz, while the IMU data is at 100 Hz. \\
\textbf{EuRoC Micro Aerial Vehicle dataset} \cite{euroc} It contains tightly synchronized video streams from a Micro Aerial Vehicle (MAV), carrying a stereo camera and an IMU, and is composed by 11 flight trajectories in two environments, exhibiting complex motion. We used Sequence \textit{MH\_04\_difficult} for testing, and left the other sequences for training. We downsampled the images and IMUs to 10 Hz and 100 Hz respectively.\\
\textbf{PennCOSYVIO dataset} \cite{penncosyvio} It is composed by four sequences where the user is carrying multiple visual and inertial sensors rigidly attached. 
We used Sequences \textit{bs}, \textit{as} and \textit{bf} for training, and \textit{af} for testing. The images and IMUs were downsampled to 10 Hz and 100 Hz respectively.

\subsection{Data Degradation}
In order to provide an extensive study of the effects of sensor data degradation and to evaluate the performances of the proposed approach, we generate three categories of degraded datasets, by adding various types of noise and occlusion to the original data, as described in the following subsections.

\vspace{-0.3cm}
\subsubsection{Vision Degradation}
\textbf{Occlusions:} we overlay a mask of dimensions 128$\times$128 pixels on top of the sample images, at random locations for each sample. Occlusions can happen due to dust or dirt on the sensor or stationary objects close to the sensor \cite{Wang2015}.\\ 
\textbf{Blur+noise:} we apply Gaussian blur with $\sigma$=15 pixels to the input images, with additional salt-and-pepper noise. Motion blur and noise can happen when the camera or the light condition changes substantially \cite{Couzinie-Devy2013}.\\
\textbf{Missing data:} we randomly remove 10\% of the input images. This can occur when packets are dropped from the bus due to excess load or temporary sensor disconnection. It can also occur if we pass through an area of very poor illumination e.g. a tunnel or underpass.


\vspace{-0.3cm}
\subsubsection{IMU Degradation}
\textbf{Noise+bias:} on top of the already noisy sensor data we add additive white noise to the accelerometer data and a fixed bias on the gyroscope data. This can occur due to increased sensor temperature and mechanical shocks, causing inevitable thermo-mechanical white noise and random walking noise \cite{NaserEl-SheimyHaiyingHou2008}.
\\
\textbf{Missing data:} we randomly remove windows of inertial samples between two consecutive random visual frames. This can occur when the IMU measuring is unstable or packets are dropped from the bus. 
\vspace{-0.3cm}
\subsubsection{Cross-Sensor Degradation}
\textbf{Spatial misalignment:} we randomly alter the relative rotation between the camera and the IMU, compared to the initial extrinsic calibration. This can occur due to axis misalignment and the incorrect sensor calibration \cite{Li2013b}. We uniformly model up to 10 degrees of misalignment .
\\
\textbf{Temporal misalignment:} we apply a time shift between windows of input images and windows of inertial measurements. This can happen due to relative drifts in clocks between independent sensor subsystems \cite{Ling2018}.

\subsection{Detailed Investigation on Robustness to Data Corruption}
Table \ref{tab:kittiseparatestudy} shows the relative performance of the proposed data fusion strategies, compared with the baselines.
In particular, we compare with a DeepVO \cite{deepvo} (Vision-Only) implementation, and finally with an implementation of VINet \cite{clarkwang2017} (VIO Direct), which uses a na\"ive fusion strategy by concatenating visual and inertial features. 
Figure \ref{fig:trajectories} shows a visual comparison of the resulting test trajectories in presence of visual and combined degradations. 
In the vision degraded set the input images are randomly degraded by adding occlusion, blurring+noise and removing images, with 10\% probability for each degradation. In the full degradation set,  images and IMU sequences from the dataset are corrupted by all seven degradations with a probability of 5\% each.
As a metric, we always report the average absolute error on relative translation and rotation estimates over the trajectory, 
in order to avoid the shortcomings of approaches using global reference frames to compute errors.

Some interesting behaviours emerge from Table \ref{tab:kittiseparatestudy}. Firstly, as expected, both the proposed fusion approaches outperform VO and the baseline VIO fusion approaches when subject to degradation.
Our intuition is that the visual features are likely to be local and discrete, and as such, erroneous regions can be blanked out, which would benefit the fusion network when it is predominantly relying on vision. Conversely, inertial data is continuous and thus a more gradual reweighting as performed by the soft fusion approach would preserve these features better. As inertial data is more important for rotation, this could explain this observation.
More interestingly, the soft fusion always improves the angle component estimation, while the hard fusion always improves the translation component estimation.
	
	\begin{table}[t]
  		\caption{Results on autonomous driving scenario \cite{Geiger2013}.} 
  		\vspace{-0.2cm}
  		\label{tab:kitti}
  		\centering
  		\resizebox{\columnwidth}{!}{%
  		\begin{tabular}{c|c|c|c}
    		 &Normal Data  & Vision Degradation & All Degradation \\
    		\hline
    		Vision Only&0.116,0.136& 0.177,0.355& \textbf{0.142},0.281\\
            VIO Direct& 0.116,0.106 & 0.175,0.164 & 0.148,0.139 \\
            VIO Soft&0.118,\textbf{0.098}&0.173,\textbf{0.150} & 0.152,\textbf{0.134}\\
            VIO Hard&\textbf{0.112},0.110&\textbf{0.172},0.151& 0.145,0.150\\
  		\end{tabular}
  		}
	\end{table}

		\begin{table}[t]
  		\caption{Results on UAV scenario \cite{euroc}.}
  		\vspace{-0.2cm}
  		\label{tab:euroc}
  		\centering
  		\resizebox{\columnwidth}{!}{%
  		\begin{tabular}{c|c|c|c}
    		&Normal Data       & Vision Degradation          & All Degradation \\
    		\hline
    		Vision Only & 0.00976,0.0867                  & 0.0222,0.268    & 0.0190,0.213\\  
            VIO Direct  & \textbf{0.00765},\textbf{0.0540}  & 0.0181,0.0696 & 0.0162,0.0935 \\ 
            VIO Soft   &  0.00848,0.0564  & \textbf{0.0170},\textbf{0.0533}          & \textbf{0.0152},0.0860 \\ 
            VIO Hard    &  0.00795,0.0589 & 0.0177,0.0565 & 0.0157,\textbf{0.0823}\\
  		\end{tabular}
  		}
	\end{table}
	
		\begin{table}[t]
  		\caption{Results on handheld scenario \cite{penncosyvio}.} 
  		\vspace{-0.2cm}
  		\label{tab:penn}
  		\centering
  		 \resizebox{\columnwidth}{!}{%
  		\begin{tabular}{c|c|c|c}
    		 &Normal Data  & Vision Degradation & All Degradation \\
    		\hline
    		Vision Only &0.0379,1.755 & 0.0446,1.849& 0.0414,1.875\\
            VIO Direct& \textbf{0.0377},1.350  & \textbf{0.0396}, 1.223& 0.0407,1.353  \\
            VIO Soft& 0.0381,\textbf{1.252} &0.0399,\textbf{1.166} & 0.0405,1.296\\
            VIO Hard& 0.0387,1.296 & 0.0410,1.206& \textbf{0.0400},\textbf{1.232}\\
  		\end{tabular}
  		}
	\end{table}
	
	 \begin{table}[h]
  	\caption{Comparison with classical methods}
  	\vspace{-0.2cm}
  	\label{tab:classical}
  	\centering
  	\resizebox{\columnwidth}{!}{%
  	\begin{tabular}{c|c|c|c|c}
		&Normal data       & Full visual degr. & Occl.+blur  & Full sensor degr.\\
		\hline
		KITTI & 0.116,0.044  & Fail & 2.4755,0.0726  & Fail\\
    	EuRoC & 0.0283,0.0402 & 0.0540,0.0591  &  0.0198,0.0400 & Fail \\ 
  	\end{tabular}
  	}
  	\vspace{-0.5cm}
    \end{table}

\subsection{Results on autonomous driving, UAV scenario and hand-held scenario}
Table \ref{tab:kitti} shows the aggregate results on the KITTI dataset in presence of normal data, all combined visual degradation and all combined visual+inertial degradation.
In particular, we compare with 
two deep approaches: DeepVO (Vision-Only) and an implementation of VINet (VIO Direct). 
We can see the same fusion behavior as in Table \ref{tab:kittiseparatestudy}.


Table \ref{tab:euroc} reports the error results on EuRoC.
Similar to KITTI, the soft fusion strategy consistently improves the angle estimation, while the hard fusion always improves the translation estimation.
Interestingly, in the hand-held scenario (Table \ref{tab:penn}) there is less marked difference between the different fusion strategies regarding the translation component.
This can be due to the small size of the dataset and the nature of motion, leading the network to slightly overfit on linear translations.
However, hard fusion still improves both errors in presence of both visual and inertial degradation. This could be ascribed to the direct fusion method overfitting on visual data, while a few transitions from outdoor to indoor introduce illumination changes and occlusion.

    \begin{figure}[t]
     	\centering
         \includegraphics[width=0.6\columnwidth]{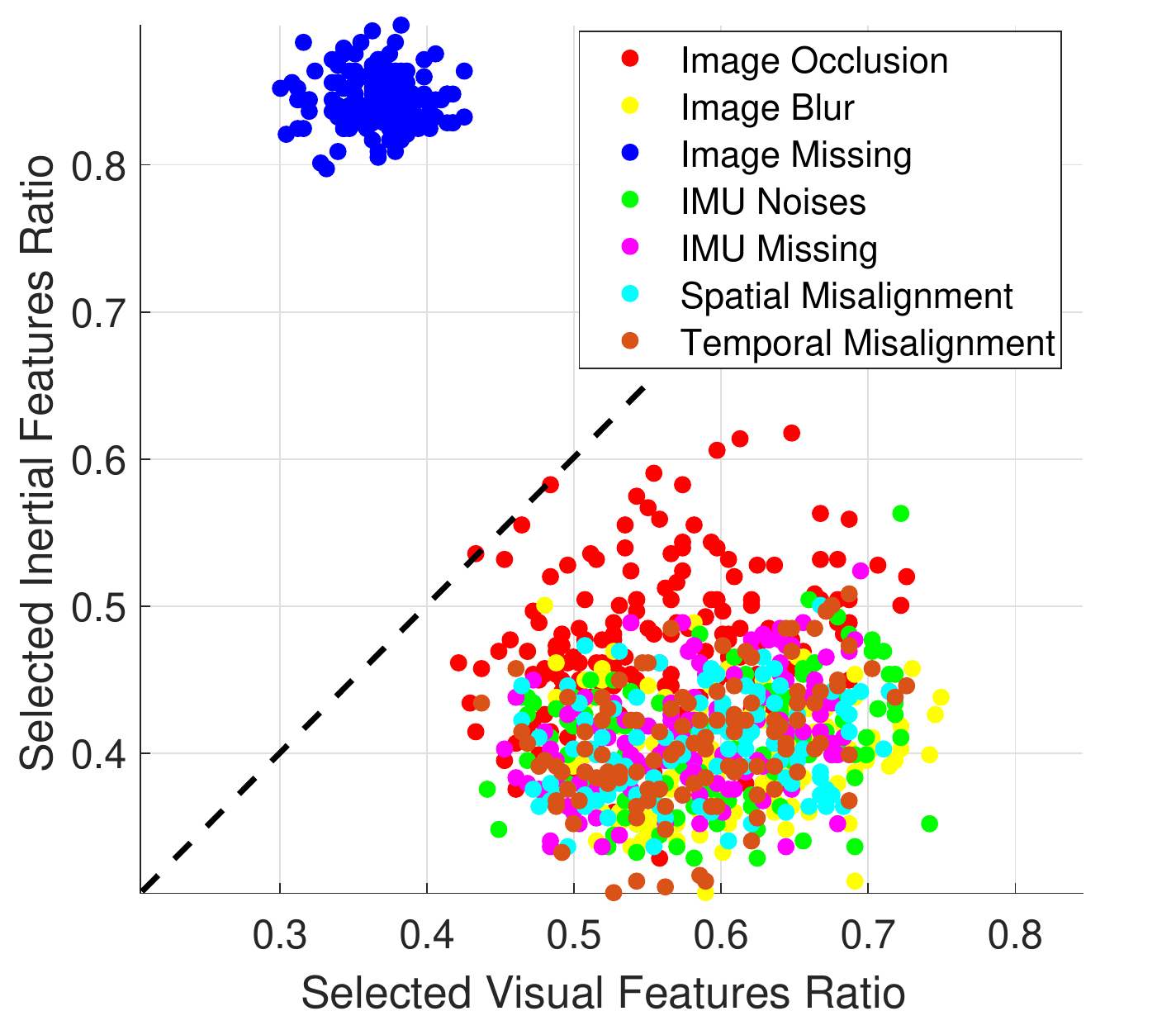}
         \caption{A comparison of visual and inertial features selection rate in seven data degradation scenarios.}
         \label{fig:interept_degration}
         \vspace{-0.3cm}
     \end{figure}

\subsection{Comparison with Classical VIOs}
For KITTI, due to the lack of time synchronization between IMUs and images, both OKVIS \cite{okvis} and VINS-Mono \cite{vinsmono} cannot work. We instead provide results from an implementation of MSCKF \cite{Hu2014} \footnote{The code can be found at: https://uk.mathworks.com/matlabcentral/\\fileexchange/43218-visual-inertial-odometry}.
For EuRoC MAV we compare with OKVIS \cite{okvis} \footnote{The code can be found at: https://github.com/ethz-asl/okvis}.

As shown in Table \ref{tab:classical}, on KITTI, MSCKF fails with full degradation due to the missing images; on EuRoc OKVIS handles missing images instead but both baselines fail with full sensor degradation due to the temporal misalignment. Learning-based methods reach comparable position/translation errors, but the orientation error is always lower for traditional methods. Because DNNs shine at extracting features and regressing translation from raw images, while IMUs improve filtering methods to get better orientation results on normal data.
Interestingly, the performance of learning-based fusion strategies degrade gracefully in the presence of corrupted data, while filtering methods fail abruptly with the presence of large sensor noise and misalignment issues.

\subsection{Interpretation of Selective Fusion} 
Incorporating hard mask into our framework enables us to quantitatively and qualitatively interpret the fusion process. 
Firstly, we analyse the contribution of each individual modality in different scenarios. Since hard fusion blocks some features according to their reliability, in order to interpret the ``feature selection'' mechanism we simply compare the ratio of the non-blocked features for each modality. 
Figure \ref{fig:interept_degration} shows that visual features dominate compared with inertial features in most scenarios. Non-blocked visual features are more than $60\%$, underlining the importance of this modality. 
We see no obvious change when facing small visual degradation, such as image blur, because the FlowNet extractor can deal with such disturbances. However, when the visual degradation becomes stronger the role of inertial features becomes significant. Notably, the two modalities contribute equally in presence of occlusion. Inertial features dominate with missing images by more than $90\%$. 

\begin{figure}[t]
    	\centering
        \begin{subfigure}[t]{0.23\textwidth}
        	\includegraphics[width=\textwidth]{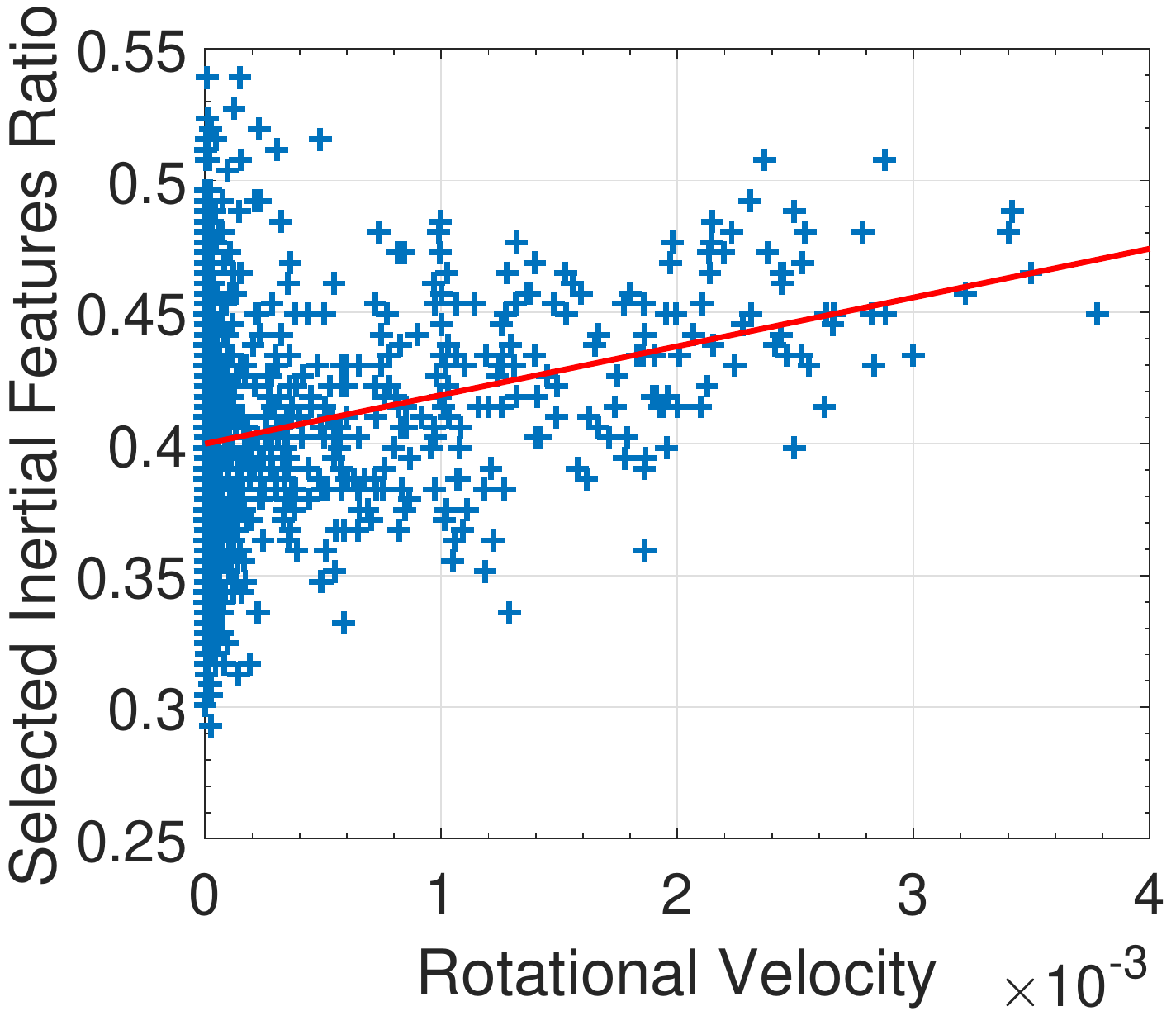}
        	\caption{\label{fig:traj_vicon} Inertial-Rotation}
        \end{subfigure}
        \begin{subfigure}[t]{0.23\textwidth}
        	\includegraphics[width=\textwidth]{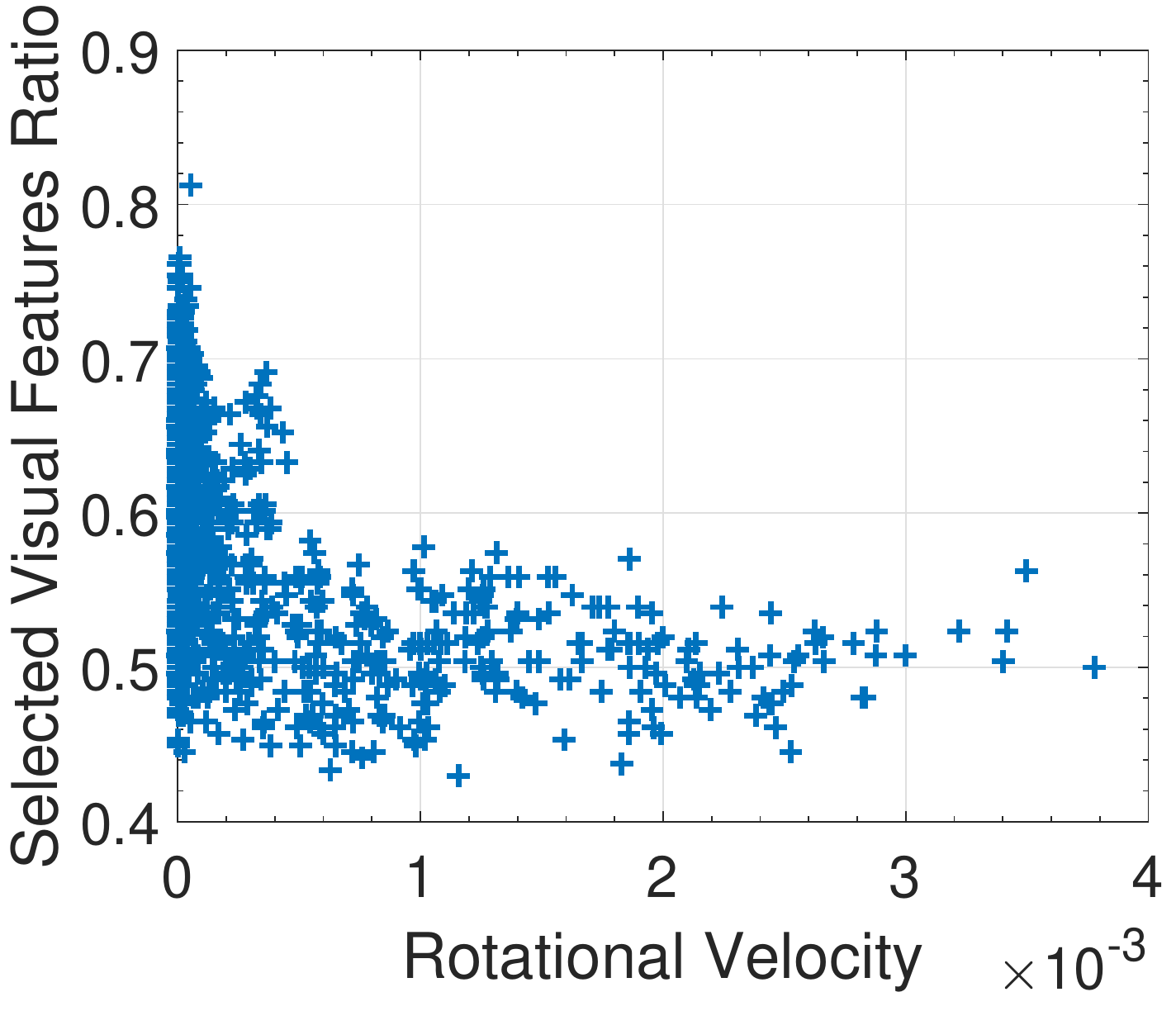}
        	\caption{\label{fig:traj_ionet} Visual-Rotation}
        \end{subfigure}
        \begin{subfigure}[t]{0.23\textwidth}
        	\includegraphics[width=\textwidth]{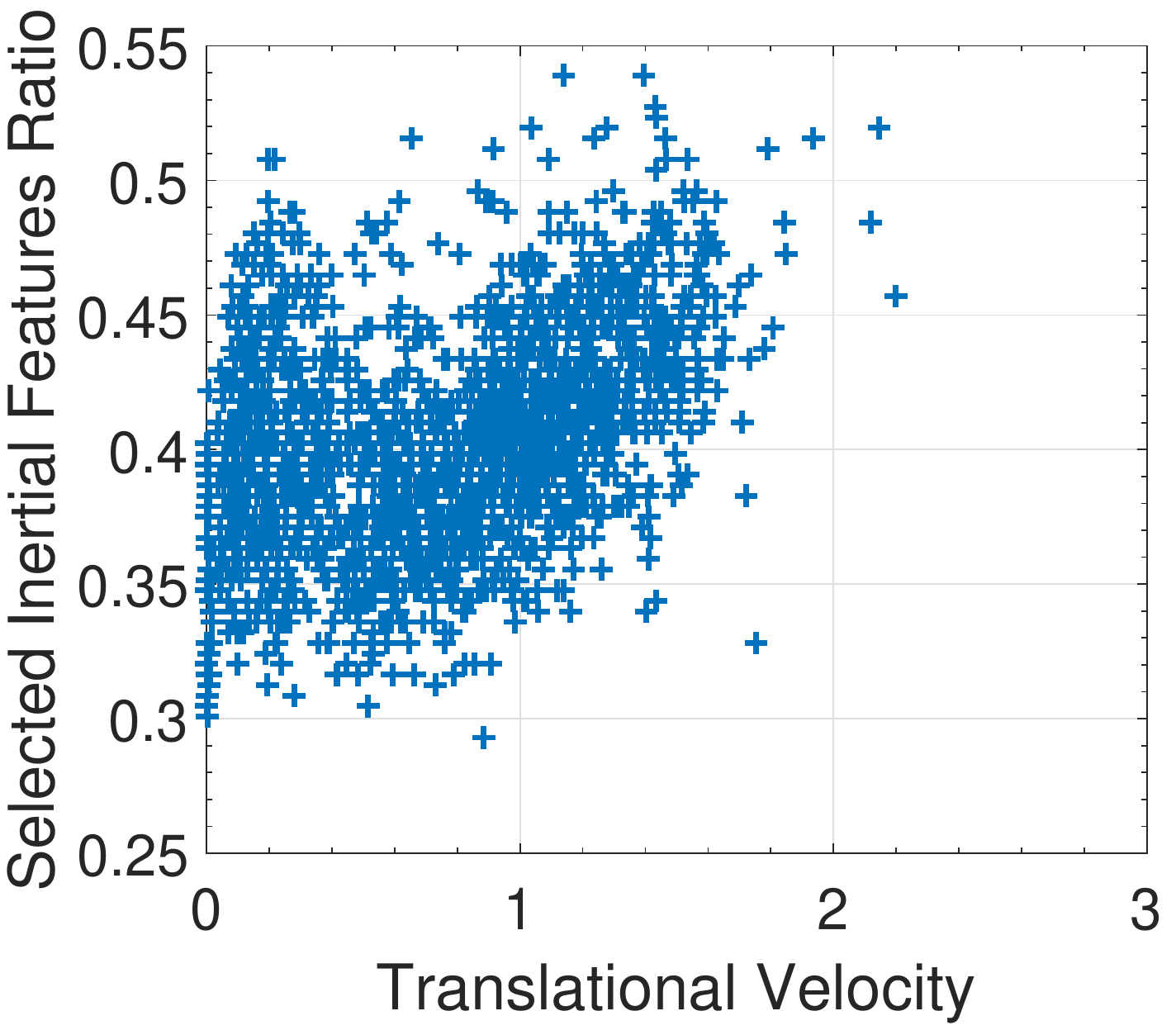}
        	\caption{\label{fig:traj_tango} Inertial-Translation}
        \end{subfigure}
         \begin{subfigure}[t]{0.23\textwidth}
        	\includegraphics[width=\textwidth]{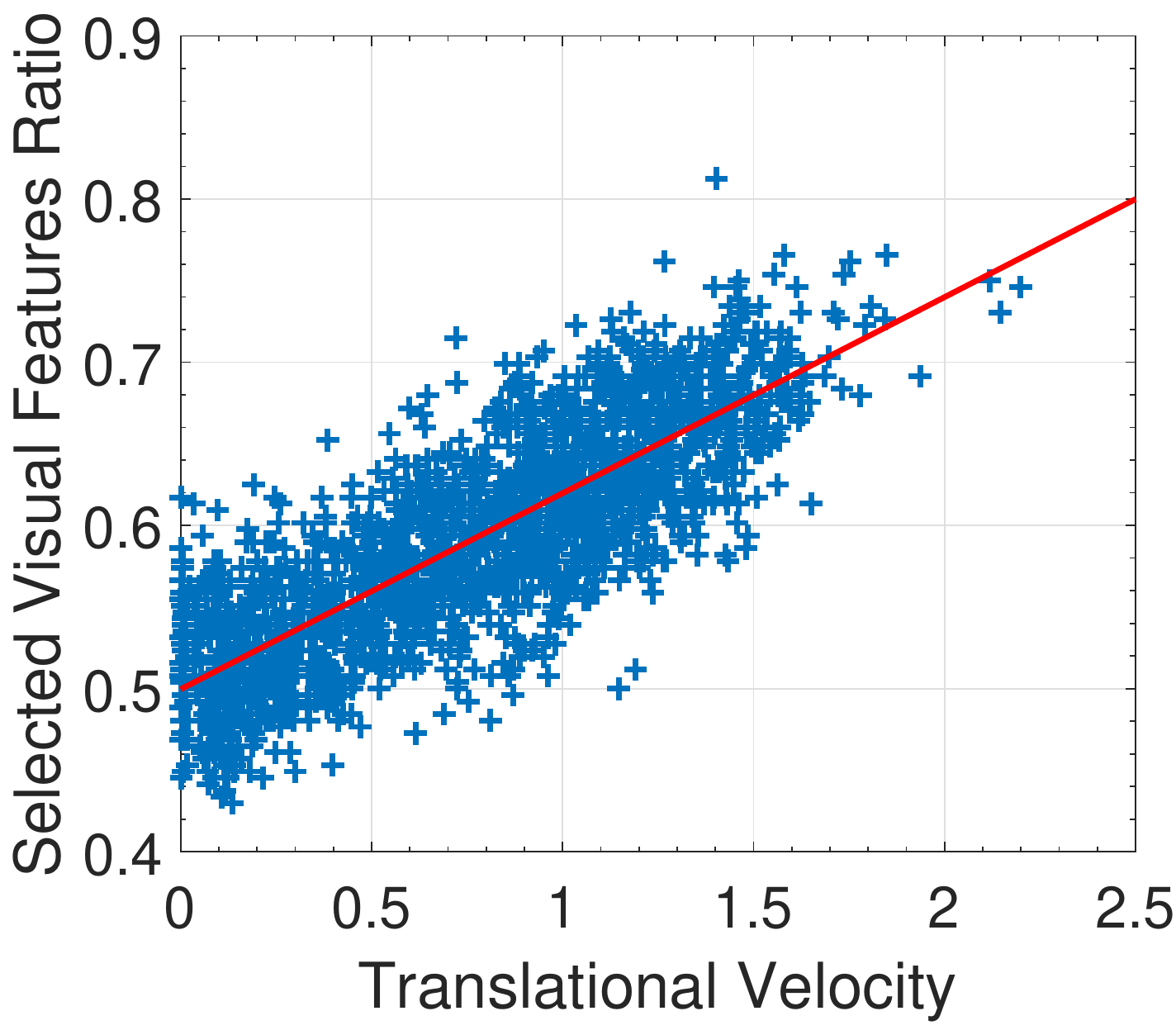}
        	\caption{\label{fig:traj_vicon} Visual-Translation}
        \end{subfigure}
        \caption{\label{fig:correlation} Correlations between the number of inertial/visual features and amount of rotation/translation show that the inertial features contribute more with rotation rates, e.g. turning, while more visual features are selected with increasing linear velocity.}
        \vspace{-0.3cm}
    \end{figure}

In Figure \ref{fig:correlation} we analyze the correlation between amount of linear and angular velocity and the selected features.
These results also show how the belief on inertial features is stronger in presence of large rotations, e.g. turning, while visual features are more reliable with increasing linear translations. It is interesting to see that at low translational velocity (0.5m / 0.1s) only 50\% to 60\% visual features are activated, while at high speed (1.5m /  0.1s) 60 \% to 75 \% visual features are used. 

\section{Related Work}
\textbf{Visual Inertial Odometry}
Traditionally, visual-inertial approaches can be roughly segmented into three different classes according to their information fusion methods:
filtering approaches \cite{jones2011visual}, fixed-lag smoothers \cite{okvis} and full smoothing methods \cite{forster2017manifold}.
In classical VIO models, their features are handcrafted, as OKVIS \cite{okvis} presented a keyframe-based approach that jointly optimizes visual feature reprojections and inertial error terms. Semi-direct \cite{tanskanen2015semi} and direct \cite{vi-dso} methods have been proposed in an effort to move towards feature-less approaches, removing the feature extraction pipeline for increased speed.
Recent VINet \cite{clarkwang2017} used neural network to learn visual-inertial navigation, but only fused two modalities in a naive concatenation way. 
We provide a generic framework for deep features fusion, and outperformed the direct fusion in different scenarios.

\textbf{Deep Neural Networks for Localization}
Recent data-driven approaches to visual odometry have gained a lot of attention. The advantage of learned methods is their potential robustness to lack of features, dynamic lightning conditions, motion blur, accurate camera calibration, which are hard to model by hard \cite{Sunderhauf2018}. Posenet \cite{kendall2015posenet} used Convolutional Neural Networks (CNNs) for 6-DoF pose regression from monocular images.
The combination of CNNs and Long-Short Term Memory (LSTM) networks was reported in \cite{Clark2017,deepvo}, showing comparable results to traditional methods. 
Several approaches \cite{zhou2017unsupervised,Yin2018,Zhan2018} used the view synthesis as unsupervisory signal to train and estimate both ego-motion and depth.
Other DL-based methods can be found on learning representations for dense visual SLAM \cite{Bloesch2018}, general map \cite{Brahmbhatt2018}, global pose \cite{Parisotto2018}, deep Localization and segmentation \cite{Wang2018}. We study the contribution of multimodal data to robust deep localization in degraded scenarios.

\textbf{Multimodal Sensor fusion and Attention}
Our proposed selective sensor fusion is related with the attention mechanisms, widely applied in neural machine translation \cite{Vaswani2017}, image caption generation \cite{Xu2015}, and video description \cite{Hori2017}. Limited by the fixed-length vector in embedding space, these attention mechanisms compute a focus map to help the decoder, when generating a sequence of words. This is different from our design intention that the features selection works to fuse multimodal sensor fusion for visual inertial odometry, and cope with more complex error resources, and self-motion dynamics.

\section{Conclusion}
In this work, we presented a novel study of end-to-end sensor fusion for visual-inertial navigation. Two feature selection strategies are proposed: deterministic soft fusion, in which a soft mask is learned from the concatenated visual and inertial features, and a stochastic hard fusion, in which Gumbel-softmax resampling is used to learn a stochastic binary mask. Based on the extensive experiments, we also provided insightful interpretations of selective sensor fusion and investigate the influence of different modalities under different degradation and self-motion circumstances.

{\small
\bibliographystyle{ieee}
\bibliography{egbib}

\begin{thebibliography}{10}\itemsep=-1pt

\bibitem{Bishop2006}
C.~Bishop.
\newblock {\em {Pattern Recognition and Machine Learning}}.
\newblock Springer, 2006.

\bibitem{rovio}
M.~Bloesch, M.~Burri, S.~Omari, M.~Hutter, and R.~Siegwart.
\newblock Iterated extended kalman filter visual-inertial odometry using direct
  photometric feedback.
\newblock {\em The International Journal of Robotics Research},
  36(10):1053--1072, 2017.

\bibitem{Bloesch2018}
M.~Bloesch, J.~Czarnowski, R.~Clark, S.~Leutenegger, and A.~J. Davison.
\newblock {CodeSLAM — Learning a Compact, Optimisable Representation for
  Dense Visual SLAM}.
\newblock In {\em CVPR}, 2018.

\bibitem{Brahmbhatt2018}
S.~Brahmbhatt, J.~Gu, K.~Kim, J.~Hays, and J.~Kautz.
\newblock {Geometry-Aware Learning of Maps for Camera Localization}.
\newblock In {\em CVPR}, pages 2616--2625, 2018.

\bibitem{euroc}
M.~Burri, J.~Nikolic, P.~Gohl, T.~Schneider, J.~Rehder, S.~Omari, M.~W.
  Achtelik, and R.~Siegwart.
\newblock The euroc micro aerial vehicle datasets.
\newblock {\em The International Journal of Robotics Research}, 2016.

\bibitem{ionet2018}
C.~Chen, C.~X. Lu, A.~Markham, and N.~Trigoni.
\newblock Ionet: Learning to cure the curse of drift in inertial odometry.
\newblock In {\em AAAI Conference on Artificial Intelligence (AAAI)}, 2018.

\bibitem{Clark2017}
R.~Clark, S.~Wang, A.~Markham, N.~Trigoni, and H.~Wen.
\newblock {VidLoc: A Deep Spatio-Temporal Model for 6-DoF Video-Clip
  Relocalization}.
\newblock In {\em CVPR}, 2017.

\bibitem{Couzinie-Devy2013}
F.~Couzinie-Devy, J.~Sun, K.~Alahari, and J.~Ponce.
\newblock {Learning to estimate and remove non-uniform image blur}.
\newblock In {\em CVPR}, pages 1075--1082, 2013.

\bibitem{Fetsch2009}
C.~R. Fetsch, A.~H. Turner, G.~C. DeAngelis, and D.~E. Angelaki.
\newblock {Dynamic Reweighting of Visual and Vestibular Cues during Self-Motion
  Perception}.
\newblock {\em Journal of Neuroscience}, 29(49):15601--15612, 2009.

\bibitem{Fischer2015}
P.~Fischer, E.~Ilg, H.~Philip, C.~Hazırbas, P.~V.~D. Smagt, D.~Cremers, and
  T.~Brox.
\newblock {FlowNet: Learning Optical Flow with Convolutional Networks}.
\newblock In {\em International Conference on Computer Vision, ICCV}, 2015.

\bibitem{forster2017manifold}
C.~Forster, L.~Carlone, F.~Dellaert, and D.~Scaramuzza.
\newblock On-manifold preintegration for real-time visual--inertial odometry.
\newblock {\em IEEE Transactions on Robotics}, 33(1):1--21, 2017.

\bibitem{Geiger2013}
A.~Geiger, P.~Lenz, C.~Stiller, and R.~Urtasun.
\newblock {Vision meets robotics: The KITTI dataset}.
\newblock {\em The International Journal of Robotics Research},
  32(11):1231--1237, 2013.

\bibitem{Gumbel1954}
E.~J. Gumbel.
\newblock {\em {Statistical theory of extreme values and some practical
  applications: a series of lectures}}.
\newblock U. S. Govt. Print. Office, 1954.

\bibitem{Hori2017}
C.~Hori, T.~Hori, T.~Y. Lee, Z.~Zhang, B.~Harsham, J.~R. Hershey, T.~K. Marks,
  and K.~Sumi.
\newblock {Attention-Based Multimodal Fusion for Video Description}.
\newblock {\em Proceedings of the IEEE International Conference on Computer
  Vision}, 2017-October:4203--4212, 2017.

\bibitem{Hu2014}
J.~S. Hu and M.~Y. Chen.
\newblock {A sliding-window visual-IMU odometer based on tri-focal tensor
  geometry}.
\newblock In {\em ICRA}, pages 3963--3968. IEEE, 2014.

\bibitem{jang2016categorical}
E.~Jang, S.~Gu, and B.~Poole.
\newblock Categorical reparameterization with gumbel-softmax.
\newblock {\em arXiv preprint arXiv:1611.01144}, 2016.

\bibitem{jones2011visual}
E.~S. Jones and S.~Soatto.
\newblock Visual-inertial navigation, mapping and localization: A scalable
  real-time causal approach.
\newblock {\em The International Journal of Robotics Research}, 30(4):407--430,
  2011.

\bibitem{kendall2015posenet}
A.~Kendall, M.~Grimes, and R.~Cipolla.
\newblock Posenet: A convolutional network for real-time 6-dof camera
  relocalization.
\newblock In {\em Proceedings of the IEEE international conference on computer
  vision}, pages 2938--2946, 2015.

\bibitem{okvis}
S.~Leutenegger, S.~Lynen, M.~Bosse, R.~Siegwart, and P.~Furgale.
\newblock Keyframe-based visual--inertial odometry using nonlinear
  optimization.
\newblock {\em The International Journal of Robotics Research}, 34(3):314--334,
  2015.

\bibitem{Li2013b}
M.~Li and A.~I. Mourikis.
\newblock {High-precision, Consistent EKF-based Visual-Inertial Odometry}.
\newblock {\em The International Journal of Robotics Research}, 32(6):690--711,
  2013.

\bibitem{Ling2018}
Y.~Ling, L.~Bao, Z.~Jie, F.~Zhu, Z.~Li, S.~Tang, Y.~Liu, W.~Liu, and T.~Zhang.
\newblock {Modeling Varying Camera-IMU Time Offset in Optimization-Based
  Visual-Inertial Odometry}.
\newblock In {\em The European Conference on Computer Vision (ECCV)}, 2018.

\bibitem{maddison2016concrete}
C.~J. Maddison, A.~Mnih, and Y.~W. Teh.
\newblock The concrete distribution: A continuous relaxation of discrete random
  variables.
\newblock {\em arXiv preprint arXiv:1611.00712}, 2016.

\bibitem{Maddison2014}
C.~J. Maddison, D.~Tarlow, and T.~Minka.
\newblock {A* Sampling}.
\newblock In {\em NIPS}, pages 1--9, 2014.

\bibitem{mnih2014neural}
A.~Mnih and K.~Gregor.
\newblock Neural variational inference and learning in belief networks.
\newblock {\em arXiv preprint arXiv:1402.0030}, 2014.

\bibitem{Mourikis2007}
A.~I. Mourikis and S.~I. Roumeliotis.
\newblock {A multi-state constraint Kalman filter for vision-aided inertial
  navigation}.
\newblock In {\em Proceedings - IEEE International Conference on Robotics and
  Automation}, pages 3565--3572, 2007.

\bibitem{NaserEl-SheimyHaiyingHou2008}
N.~{Naser, El-Sheimy; Haiying, Hou; Xiaojii}.
\newblock {Analysis and Modeling of Inertial Sensors Using Allan Variance}.
\newblock {\em IEEE Transactions on Instrumentation and Measurement},
  57(JANUARY):684--694, 2008.

\bibitem{Parisotto2018}
E.~Parisotto, D.~S. Chaplot, J.~Zhang, and R.~Salakhutdinov.
\newblock {Global Pose Estimation with an Attention-based Recurrent Network}.
\newblock In {\em CVPR}, 2018.

\bibitem{penncosyvio}
B.~Pfrommer, N.~Sanket, K.~Daniilidis, and J.~Cleveland.
\newblock Penncosyvio: {A} challenging visual inertial odometry benchmark.
\newblock In {\em 2017 {IEEE} International Conference on Robotics and
  Automation, {ICRA} 2017, Singapore, Singapore, May 29 - June 3, 2017}, pages
  3847--3854, 2017.

\bibitem{vinsmono}
T.~Qin, P.~Li, and S.~Shen.
\newblock Vins-mono: A robust and versatile monocular visual-inertial state
  estimator.
\newblock {\em IEEE Transactions on Robotics}, 34(4):1004--1020, Aug 2018.

\bibitem{clarkwang2017}
H.~W. A. M. N.~T. Ronald~Clark, Sen~Wang.
\newblock Vinet: Visual-inertial odometry as a sequence-to-sequence learning
  problem.
\newblock In {\em Proceedings of the Thirty-First AAAI Conference on Artificial
  Intelligence}. AAAI, 2017.

\bibitem{Sunderhauf2018}
N.~S{\"{u}}nderhauf, O.~Brock, W.~Scheirer, R.~Hadsell, D.~Fox, J.~Leitner,
  B.~Upcroft, P.~Abbeel, W.~Burgard, M.~Milford, and P.~Corke.
\newblock {The limits and potentials of deep learning for robotics}.
\newblock {\em International Journal of Robotics Research}, 37(4-5):405--420,
  2018.

\bibitem{tanskanen2015semi}
P.~Tanskanen, T.~Naegeli, M.~Pollefeys, and O.~Hilliges.
\newblock Semi-direct ekf-based monocular visual-inertial odometry.
\newblock In {\em Intelligent Robots and Systems (IROS), 2015 IEEE/RSJ
  International Conference on}, pages 6073--6078. IEEE, 2015.

\bibitem{Vaswani2017}
A.~Vaswani, N.~Shazeer, N.~Parmar, J.~Uszkoreit, L.~Jones, A.~N. Gomez,
  L.~Kaiser, and I.~Polosukhin.
\newblock {Attention Is All You Need}.
\newblock In {\em NIPS}, 2017.

\bibitem{vi-dso}
L.~von Stumberg, V.~Usenko, and D.~Cremers.
\newblock Direct sparse visual-inertial odometry using dynamic marginalization,
  2018.

\bibitem{Wang2018}
P.~Wang, R.~Yang, B.~Cao, W.~Xu, and Y.~Lin.
\newblock {DeLS-3D: Deep Localization and Segmentation with a 3D Semantic Map}.
\newblock In {\em CVPR}, 2018.

\bibitem{deepvo}
S.~Wang, R.~Clark, H.~Wen, and N.~Trigoni.
\newblock Deepvo: Towards end-to-end visual odometry with deep recurrent
  convolutional neural networks.
\newblock {\em International Conference on Robotics and Automation}, 2017.

\bibitem{Wang2015}
T.~C. Wang, A.~A. Efros, and R.~Ramamoorthi.
\newblock {Occlusion-aware depth estimation using light-field cameras}.
\newblock In {\em Proceedings of the IEEE International Conference on Computer
  Vision}, pages 3487--3495, 2015.

\bibitem{williams1992simple}
R.~J. Williams.
\newblock Simple statistical gradient-following algorithms for connectionist
  reinforcement learning.
\newblock {\em Machine learning}, 8(3-4):229--256, 1992.

\bibitem{Xu2015}
K.~Xu, J.~Ba, R.~Kiros, K.~Cho, A.~Courville, R.~Salakhutdinov, R.~Zemel, and
  Y.~Bengio.
\newblock {Show, Attend and Tell: Neural Image Caption Generation with Visual
  Attention}.
\newblock In {\em ICML}, 2015.

\bibitem{Yang2018}
N.~Yang, R.~Wang, X.~Gao, and D.~Cremers.
\newblock {Challenges in Monocular Visual Odometry: Photometric Calibration,
  Motion Bias and Rolling Shutter Effect}.
\newblock {\em IEEE ROBOTICS AND AUTOMATION LETTERS}, pages 1--8, 2018.

\bibitem{Yin2018}
Z.~Yin and J.~Shi.
\newblock {GeoNet: Unsupervised Learning of Dense Depth, Optical Flow and
  Camera Pose}.
\newblock In {\em CVPR}, 2018.

\bibitem{Zhan2018}
H.~Zhan, R.~Garg, C.~S. Weerasekera, K.~Li, H.~Agarwal, and I.~Reid.
\newblock {Unsupervised Learning of Monocular Depth Estimation and Visual
  Odometry with Deep Feature Reconstruction}.
\newblock In {\em CVPR}, pages 340--349, 2018.

\bibitem{zhou2017unsupervised}
T.~Zhou, M.~Brown, N.~Snavely, and D.~G. Lowe.
\newblock Unsupervised learning of depth and ego-motion from video.
\newblock In {\em CVPR}, volume~2, page~7, 2017.

\end{thebibliography}
}

\clearpage


\section*{Supplementary material}

# Supplementary Material

## Abstract

*In this supplementary document we provide additional implementation details, including the detailed network architecture, and data preparation/training details. We also provide additional results and discussion on predicted global trajectories.*

## 1. Network Architecture

The implementation details for the architecture that was used in our experiments are reported in Table 1. We use the following abbreviations:
**Conv**: convolution, **FC**: fully-connected, **activ.**: activation function, $\oplus$: concatenation; $\odot$: element-wise multiplication; $B$: batch size.

The visual encoder follows the encoder structure of the FlowNetS architecture [3]. We initialize the visual encoder weights with the weights of a model that was pre-trained on the FlyingChairs dataset[1], since training from scratch would require larger amounts of data compared to our dataset sizes. For this reason, training the encoder from scratch experimentally gave worse results.

Bi-directional LSTMs were used for the inertial encoder (as in [2]) as well as in the recurrent part of the pose regressor. Bi-directional RNNs provide higher data efficiency when dealing with small datasets. The LSTM layers include dropout regularization on the recurrent connections.

## 2. Dataset Preprocessing Details

Few publicly available datasets provide camera images, high-frequency IMU and ground-truth for training visual inertial odometry (i.e., inertial data is missing in the Oxford Robotcar Dataset [5] and 7-Scenes Dataset [9]; ground-truth for the full trajectories is not available in TUM VIO [8]). Therefore, we use the KITTI [4], EuRoC [1], and PennCOSYVIO [7] datasets for training and evaluation.

**KITTI** High-frequency inertial data (100Hz) is only available in the raw unsynced data packages. Hence, we manually synchronize inertial data and images according to their timestamps. We used Sequences *00, 01, 02, 04, 06, 08, 09* for training and tested the network on Sequences *05*, *07*, and *10*, excluding sequence 03 as the corresponding raw file is unavailable. Correspondences between the KITTI Odometry Number, the raw data file names and the corresponding number of sequences are reported in Table 2. The images are resized to 512×256.

**EuRoC** For the EuRoC MAV dataset [1] we use grayscale video images from CAM1 at 20fps of the VI-Sensor and the IMU measurements from the same sensors at 200Hz. The data is tightly synchronized. Doubling the sensor rates compared to the ones in KITTI helps to deal with the faster, jerky MAV movements. As with KITTI, images are resized to 512×256. We train on all sequences, minus *MH_04_difficult*, which is used for testing.

**PennCOSYVIO** A number of sensors are included in the PennCOSYVIO dataset [7]. For our experiment we select the video images from the Tango Bottom camera, running at 30fps, and inertial measurements from the VI-Sensor, running at 200Hz. Images are subsampled to 10Hz, and IMU measurements are subsampled to 100Hz, similarly to KITTI. Images are cropped and resized to 512×256. We train on sequences *as, bf, bs* and test on sequence *af*. Small time-synchronization errors between the Tango camera and the VI-Sensor are likely present in the data, leading to worse results in all scenarios.

## 3. Evaluation of global trajectories on KITTI

Figure 1 shows the global RMSE position errors on the three test KITTI trajectories (Seq 05, Seq 07, Seq 10) as a function of travelled distance, for both the normal dataset and the fully degraded dataset (visual degradation + sensor degradation). We compare the two VO and vanilla VIO baselines with the proposed soft and hard fusion strategies. It can be noticed how, while at start VIO performs as well as soft and hard fusion, on average, over time the proposed selective fusion strategies outperform the vanilla fusion, since the increased robustness reduces error accumulation. This is particularly visible in the most challenging Seq 05. As expected, a VO approach heavily underperforms in presence of large amounts of angular rotations (Seq 05, Seq 07).

Another interesting result is how VO performs slightly better in presence of IMU degradation and camera-IMU

---

[1] https://lmb.informatik.uni-freiburg.de/resources/datasets/FlyingChairs.en.html

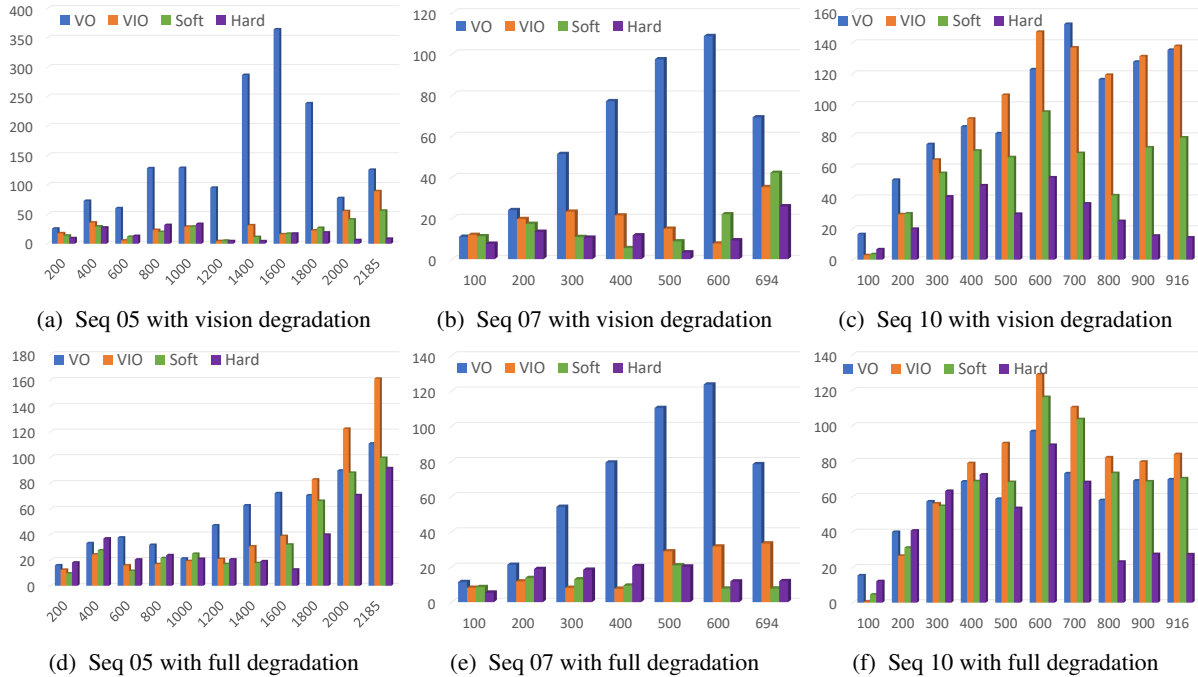

Figure 1: Global position errors (in meters, Y axis) on the KITTI dataset, over travelled distance (in meters, X axis).

synchronization (Figure 1, bottom row). That shows how a vanilla VIO fusion is unable to deal with these issues, to the point of underperforming compared to vision-only approaches. This result further corroborates the fact that in deep learning-based approaches explicitly learning the belief on the different components makes the estimation more robust, while stacking sensors without a sensible fusion strategy can lead to *catastrophic fusion*, similarly to traditional approaches. Catastrophic fusion happens when the single components of the system before fusion significantly outperform the overall system after fusion [6].

## 4. Supplementary Video

The supplementary video shows an evaluation of soft fusion and hard fusion on Seq 05 of the KITTI dataset, with vision degradation (10% occlusion, 10% blur+noise, 10% missing data), compared with two deep VO and VIO baselines. The degraded images and corresponding soft/hard masks are shown in the top-left and bottom-left respectively. The trajectories from the four methods are shown on the right side.

## References

[1] M. Burri, J. Nikolic, P. Gohl, T. Schneider, J. Rehder, S. Omari, M. W. Achtelik, and R. Siegwart. The euroc micro aerial vehicle datasets. *The International Journal of Robotics Research*, 2016. 1

[2] C. Chen, C. X. Lu, A. Markham, and N. Trigoni. Ionet: Learning to cure the curse of drift in inertial odometry. In *AAAI Conference on Artificial Intelligence (AAAI)*, 2018. 1

[3] P. Fischer, E. Ilg, H. Philip, C. Hazrbas, P. V. D. Smagt, D. Cremers, and T. Brox. FlowNet: Learning Optical Flow with Convolutional Networks. In *International Conference on Computer Vision, ICCV*, 2015. 1

[4] A. Geiger, P. Lenz, C. Stiller, and R. Urtasun. Vision meets robotics: The KITTI dataset. *The International Journal of Robotics Research*, 32(11):1231–1237, 2013. 1

[5] W. Maddern, G. Pascoe, C. Linegar, and P. Newman. 1 Year, 1000km: The Oxford RobotCar Dataset. *The International Journal of Robotics Research (IJRR)*, 36(1):3–15, 2016. 1

[6] J. R. Movellan and P. Mineiro. Robust sensor fusion: Analysis and application to audio visual speech recognition. *Machine Learning*, 32(2):85–100, 1998. 2

[7] B. Pfrommer, N. Sanket, K. Daniilidis, and J. Cleveland. Pennscosyvio: A challenging visual inertial odometry benchmark. In *2017 IEEE International Conference on Robotics and Automation, ICRA 2017, Singapore, Singapore, May 29 - June 3, 2017*, pages 3847–3854, 2017. 1

[8] D. Schubert, T. Goll, N. Demmel, V. Usenko, J. Stückler, and D. Cremers. The TUM VI Benchmark for Evaluating Visual-Inertial Odometry. In *ICRA*, 2018. 1

[9] J. Shotton, B. Glocker, C. Zach, S. Izadi, A. Criminisi, and A. Fitzgibbon. Scene coordinate regression forests for camera relocalization in RGB-D images. In *CVPR*, pages 2930–2937, 2013. 1

2

Visual Encoder

| | |
|---|---|
| [ input ] | Two stacked images: $B \times 512 \times 256 \times 6$ |
| [ layer 1 ] | Conv. $7^2$, Stride $2^2$, Padding 3, LeakyReLU activ. |
| [ layer 2 ] | Conv. $5^2$, Stride $2^2$, Padding 2, LeakyReLU activ. |
| [ layer 3 ] | Conv. $5^2$, Stride $2^2$, Padding 2, LeakyReLU activ. |
| [ layer 3_1 ] | Conv. $3^2$, Stride $1^2$, Padding 1 |
| [ layer 4 ] | Conv. $3^2$, Stride $2^2$, Padding 2, LeakyReLU activ. |
| [ layer 4_1 ] | Conv. $3^2$, Stride $1^2$, Padding 1 |
| [ layer 5 ] | Conv. $3^2$, Stride $2^2$, Padding 2, LeakyReLU activ. |
| [ layer 5_1 ] | Conv. $3^2$, Stride $1^2$, Padding 1 |
| [ layer 6 ] | Conv. $3^2$, Stride $2^2$, Padding 1 |
| [ layer 6 output ] | $B \times 4 \times 8 \times 1024$ |
| [ layer 7 ] | FC 256 |
| [ output ] | $B \times 256$ |

Inertial Encoder

| | |
|---|---|
| [ input ] | IMU sequence: $B \times 10 \times 6$ |
| [ layer 1 ] | FC 128 |
| [ layer 2 ] | B-LSTM, 2-layers, hidden size 128, Dropout 0.2 |
| [ output ] | $B \times 256$ |

Direct Fusion Module

| | |
|---|---|
| [ input ] | $B \times (256 \oplus 256)$ |
| [ output ] | $B \times 512$ |

Soft Fusion Module

| | |
|---|---|
| [ input ] | $B \times (256 \oplus 256)$ |
| [ layer 1 ] | FC 512 (input) |
| [ layer 2 ] | [ (input) $\odot$ (layer 1) |
| [ output ] | $B \times 512$ |

Hard Fusion Module

| | |
|---|---|
| [ input ] | $B \times (256 \oplus 256)$ |
| [ layer 1 ] | FC 1024, Sigmoid activ. (input) |
| [ layer 2 ] | Gumbel-Softmax sampling, $512 \times 2$ |
| [ layer 2 ] | [ (input) $\odot$ (layer 2) |
| [ output ] | $B \times 512$ |

Pose Regressor

| | |
|---|---|
| [ input ] | $B \times 512$ |
| [ layer 1 ] | B-LSTM, 2-layers, hidden size 512, Dropout 0.2 |
| [ layer 2 ] | Dropout 0.2 |
| [ layer 3_1 ] | FC 3 (layer 2) |
| [ layer 3_2 ] | FC 3 (layer 2) |
| [ layer 4 ] | (layer 3_1) $\oplus$ (layer 3_2) |
| [ output ] | $B \times 6$ |

Table 1: Implementation details for the proposed architecture. $\oplus$ denotes a concatenation operation; $\odot$ indicates an element-wise product between two tensors.

Table 2: Correspondences between the KITTI Odometry Number, the raw data file names and the corresponding number of sequences

| Odometry Nr. | Raw Data Filename | Number of Sequences |
|---|---|---|
| 00 | 2011_10_03_drive_0027 | 4390 |
| 01 | 2011_10_03_drive_0042 | 1173 |
| 02 | 2011_10_03_drive_0034 | 4485 |
| 03 | 2011_09_26_drive_0067 | Not available |
| 04 | 2011_09_30_drive_0016 | 284 |
| 05 | 2011_09_30_drive_0018 | 2750 |
| 06 | 2011_09_30_drive_0020 | 1091 |
| 07 | 2011_09_30_drive_0027 | 1111 |
| 08 | 2011_09_30_drive_0028 | 5149 |
| 09 | 2011_09_30_drive_0033 | 1599 |
| 10 | 2011_09_30_drive_0034 | 1215 |


\end{document}